\documentclass{article}
\usepackage[T1]{fontenc}
\PassOptionsToPackage{table}{xcolor}
\usepackage{iclr2027_conference,times}
\usepackage{amsmath,amssymb}
\usepackage{newtxmath}
\usepackage{graphicx,booktabs,tabularx,longtable,float,adjustbox}
\usepackage{caption}
\usepackage{xurl,hyperref}
\usepackage{xspace}
\usepackage{placeins}
\usepackage{wrapfig}
\definecolor{cordTeal}{HTML}{0796A0}
\definecolor{cordOrange}{HTML}{F28E2B}
\definecolor{cordGray}{HTML}{667482}
\definecolor{cordBest}{HTML}{D62728}
\definecolor{cordSecond}{HTML}{1F77B4}
\definecolor{cordCite}{HTML}{1769D2}
\definecolor{cordHeader}{HTML}{F1F5F8}
\definecolor{cordRow}{HTML}{FAFCFD}
\definecolor{cordTealLight}{HTML}{EAF7F6}
\hypersetup{colorlinks=true,citecolor=cordCite,urlcolor=cordCite,linkcolor=cordCite}
\newcommand{\best}[1]{\ifmmode\textcolor{cordBest}{\mathbf{#1}}\else\textbf{\textcolor{cordBest}{#1}}\fi}
\newcommand{\second}[1]{\ifmmode\textcolor{cordSecond}{\underline{\mathbf{#1}}}\else\textbf{\underline{\textcolor{cordSecond}{#1}}}\fi}
\newcommand{\lead}[1]{\noindent\textbf{#1}\hspace{0.35em}}

\newcommand{\CORD}{\textsc{CORD}\xspace}
\newcommand{\CordScratch}{\CORD (Scratch)\xspace}
\newcommand{\CordSingle}{\CORD (Single-domain)\xspace}
\newcommand{\CordMulti}{\CORD (Multi-domain)\xspace}
\newcommand{\Frozen}{Frozen\xspace}
\newcommand{\PartialFT}{Partial FT\xspace}
\newcommand{\FullFT}{Full FT\xspace}
\newcommand{\maintablefont}{\small}
\title{CORD: Learning Reusable Degradation Representations Across Heterogeneous Physical Systems}
\author{Haibo Li$^{1}$, Zhiguo Zeng$^{1}$\\
  {\normalfont\footnotesize $^{1}$CentraleSup\'elec, Universit\'e Paris-Saclay}\\
  {\normalfont\footnotesize haibo.li@centralesupelec.fr\quad zhiguo.zeng@centralesupelec.fr}}
\begin{document}
\maketitle
\begin{abstract}
Can heterogeneous physical degradation systems benefit from joint pretraining
and move beyond system-specific prognostics toward reusable cross-system
representation learning? CORD combines type-specific observation interfaces
with a shared degradation backbone. Its two self-supervised objectives learn
at complementary scales: Intra-Observation Structure Modeling (ISM) captures
structure within observations, while Inter-Observation Dynamics Modeling (IDM)
captures latent degradation evolution across observation histories. We evaluate
CORD under two transfer boundaries: Pretraining-Included System Types, where
downstream datasets and held-out units are unseen but their system types are
represented during source pretraining, and Pretraining-Excluded System Types,
where the entire turbofan-engine type is absent from pretraining. Across
bearings, batteries, and cutting tools, CORD (Multi-domain) consistently
improves over CORD (Single-domain) under Frozen adaptation, provides further
gains under Full FT in most settings, and remains competitive with
representative external baselines. Source-pretrained initialization also
improves low-label adaptation to the pretraining-excluded engine type.
Frozen-representation analysis further shows improved cross-unit lifecycle
consistency after multi-domain pretraining. Joint pretraining across
heterogeneous physical systems thus produces degradation representations
reusable across devices, datasets, and system types.
\end{abstract}

\section{Introduction}

Physical degradation is observed through system-dependent measurements: bearings through vibration, batteries through electrochemical cycling, cutting tools through machining signals, and turbofan engines through multivariate flight histories. These observations differ in channel semantics, physical units, sampling cadence, and operating context. Industrial prognostics is therefore commonly developed in a system-specific manner, with each model learning degradation patterns from the measurements available for one asset type. Transfer-learning approaches can reduce this isolation by adapting knowledge across operating conditions or related prognostic domains \citep{dacosta2020rul,wang2026heterogeneous}; however, they do not directly answer whether degradation knowledge can be learned jointly across physically different systems whose observations are not semantically aligned.

Large-scale pretraining has produced reusable temporal representations across diverse time-series datasets. MOMENT studies general-purpose representations and limited-supervision adaptation, while Moirai and Time-MoE address heterogeneous forecasting through universal models and sparse mixture-of-experts pretraining \citep{goswami2024moment,woo2024moirai,shi2025timemoe}. Tabular foundation models offer another route to data-efficient PHM prediction \citep{theiler2026phm}. FeDaL addresses dataset-level heterogeneity in time-series pretraining, and FORMED adapts a shared backbone to medical datasets with different channel structures and tasks \citep{chen2026fedal,huang2026formed}. Existing work has not established whether physically distinct degradation systems can contribute to a shared representation learner when their sensing semantics are not aligned. Section~2 reviews these directions alongside prognostics transfer.

\textbf{CORD} addresses this problem by keeping observation interfaces type-specific while sharing the degradation backbone across systems. Each interface preserves its system's sensing semantics while mapping observations into the shared backbone. Two self-supervised objectives train the encoder. \textbf{Intra-Observation Structure Modeling (ISM)} captures structural relationships within a health-state observation, whereas \textbf{Inter-Observation Dynamics Modeling (IDM)} captures how latent health states evolve across observation histories. Lightweight target-aware prediction heads reuse the resulting representation, allowing the encoder to be evaluated independently of any single downstream RUL architecture.

We evaluate CORD under two transfer boundaries. \textbf{Level I: Generalization under Pretraining-Included System Types} evaluates downstream datasets that are entirely excluded from source pretraining while other datasets from the same bearing, battery, or cutting-tool type are available upstream. \textbf{Level II: Generalization under Pretraining-Excluded System Types} removes the complete target system type from source pretraining and introduces it only during downstream adaptation. XJTU-SY bearings \citep{wang2020hybrid}, CALCE CS2 batteries \citep{calceBatteryData}, PHM2010 cutting tools \citep{phmsociety2010challenge}, and N-CMAPSS turbofan engines \citep{ariaschao2021ncmapss} instantiate these two boundaries. Across Level I, CORD (Multi-domain) improves over CORD (Single-domain) in all nine Frozen settings and eight of nine Full-FT settings, while frozen-representation retrieval error decreases by 13.9\%, 17.9\%, and 27.3\% for bearings, batteries, and cutting tools, respectively. At Level II, source pretraining also improves low-label adaptation to the previously unseen turbofan-engine type. These results indicate that the learned degradation representation is reusable beyond individual datasets and devices.

Our contributions are:
\begin{enumerate}
    \item \textbf{Cross-system degradation representation architecture.}
    CORD combines type-specific observation interfaces with a shared degradation
    backbone, preserving native sensing semantics across physically heterogeneous
    systems while learning a shared degradation representation.

    \item \textbf{Complementary structure--dynamics pretraining.}
    CORD learns degradation information at two complementary scales: ISM models
    structural dependencies within an observation, while IDM models latent state
    evolution across observation histories.

    \item \textbf{Reusable multi-domain pretraining across system boundaries.}
    Multi-domain pretraining improves representation reuse on unseen datasets and
    supports low-label adaptation to a physical system type entirely excluded from
    source pretraining.
\end{enumerate}

\section{Related Work}
\lead{Time-series foundation models.} Large-scale pretraining has produced
reusable models across heterogeneous time-series corpora. Forecasting models
address variation in frequency, variate structure, and distribution through
universal, decoder-based, or generative pretraining
\citep{woo2024moirai,das2024timesfm,ansari2024chronos,liu2024timer,shi2025timemoe}.
General-purpose and representation-oriented models also learn from temporal
structure beyond forecasting targets, including limited-supervision
adaptation, time-aware representations, subseries dependencies, and joint
reconstruction and autoregression
\citep{goswami2024moment,fraikin2024trep,dong2024timesiam,he2026gtm}.
CORD focuses on physical degradation systems whose sensors and measured
variables may have different meanings. It retains type-specific observation
interfaces and shares representation learning above them.

\lead{Transfer and representation learning for prognostics.} RUL transfer
methods address operating-condition shifts and heterogeneous feature spaces
\citep{dacosta2020rul,wang2026heterogeneous}; tabular foundation models
provide another interface for heterogeneous PHM tasks
\citep{theiler2026phm}. CORD learns source representations jointly from
multiple physical system types and tests their reuse on held-out datasets and
on a target type absent from source pretraining.

\lead{Learning under heterogeneous observations.} Residual and transfer
adapters combine shared transformations with specialized responses
\citep{rebuffi2017adapters,houlsby2019adapter}. FeDaL handles dataset-specific
heterogeneity in federated pretraining, while FORMED adapts foundation models
across medical observation interfaces
\citep{chen2026fedal,huang2026formed}. In CORD, type-specific interfaces
preserve the native measurements of each degradation system during centralized
self-supervised pretraining of a shared backbone.

\section{CORD: Reusable Degradation Representation Learning}\label{sec:method}

\subsection{Problem setup and transfer boundaries}\label{sec:problem}
Let $d$ denote a physical system type, $k$ a dataset, $i$ an individual unit,
and $t$ an ordered observation. A physical system type is a broad category of
degrading assets, such as bearings, batteries, cutting tools, or turbofan
engines. Each type can contain multiple datasets collected from different
units, operating conditions, sensors, and acquisition protocols. The hierarchy
is therefore system type $\rightarrow$ dataset $\rightarrow$ unit
$\rightarrow$ observation. In the names of source-pretraining regimes,
\emph{domain} denotes a physical system type. A time-indexed sample is
represented as a health-state observation $\mathcal O_{i,t}^{d,k}$. Source
pretraining uses system-type set $\mathcal D_{\mathrm{pre}}$ and dataset
collection $\mathcal P$. Shared parameters $\theta$ and type-specific
parameters $\phi_d$ produce
$e_{i,t}=E_{\theta,\phi_d}(\mathcal O_{i,t}^{d,k})$.
A target-aware readout maps an observation history to scalar remaining useful
life (RUL). The primary learning objective is a reusable degradation
representation; RUL prediction provides downstream tests of its reuse.

\textbf{Level I: Generalization under Pretraining-Included System Types.}
Other datasets from the target system type belong to the source-pretraining
collection $\mathcal P$, but the entire downstream target dataset is excluded;
training and test units in that dataset are disjoint.
\textbf{Level II: Generalization under Pretraining-Excluded System Types.}
The target system type $d^\star$ and all its datasets are excluded from source
pretraining. A new interface and prediction model may subsequently be trained
using labeled observations from training units; test units do not enter model
fitting or protocol selection. Both levels evaluate reuse from the same
source-pretrained CORD model under different target boundaries. \Frozen
adaptation tests what can be read out without changing the encoder; \FullFT
tests the value of its initialization.

\begin{figure}[t]
\centering\includegraphics[width=.98\linewidth]{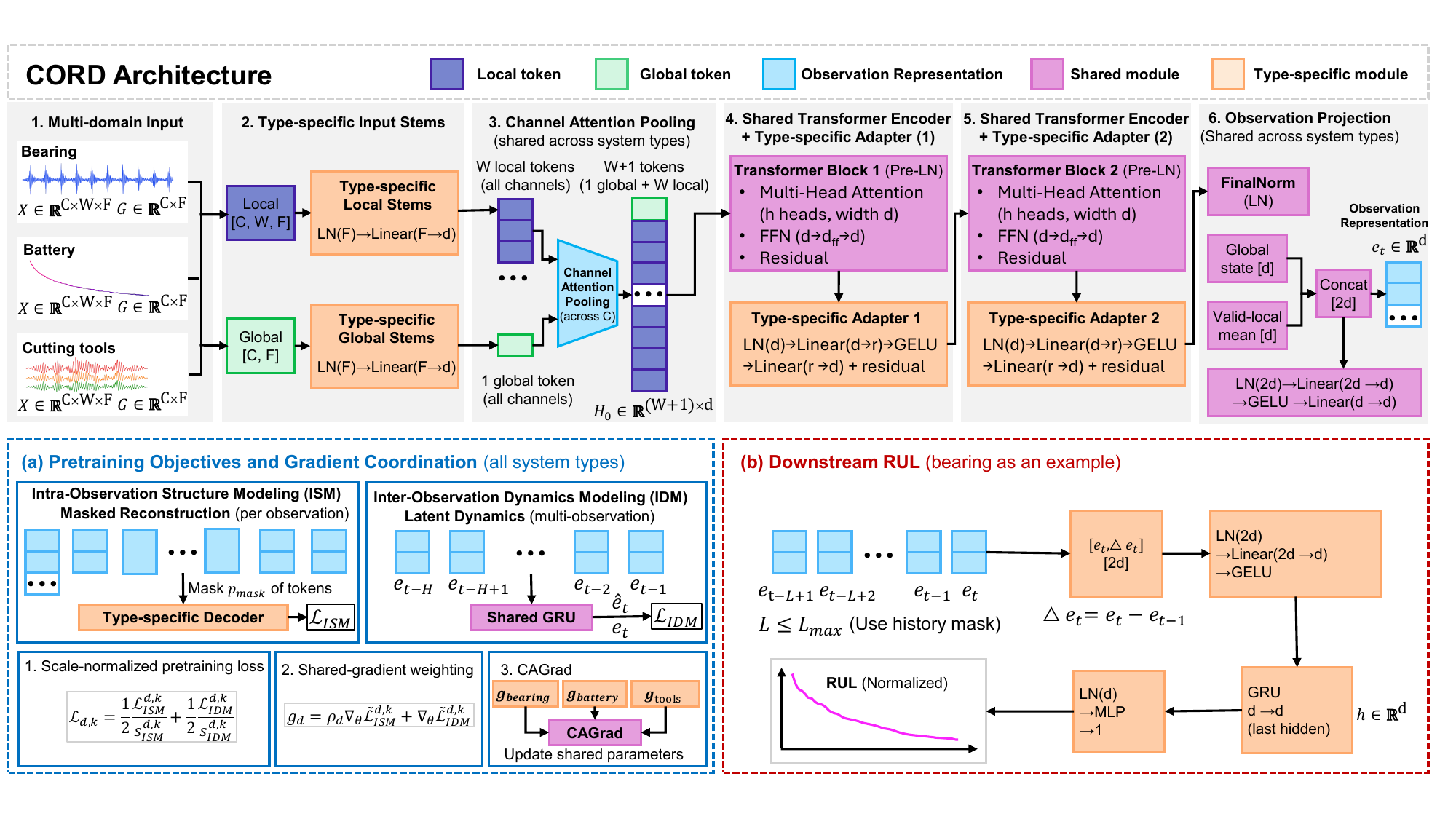}
\caption{CORD framework. Type-specific observation interfaces preserve
heterogeneous measurement semantics, while a shared Transformer backbone
learns reusable degradation representations. ISM models structure within an
observation and IDM models evolution across observation histories; target-aware
heads reuse the encoder for downstream prognostics. Symbolic dimensions are
instantiated in Appendices~\ref{app:descriptors} and~\ref{app:optimization}.}
\label{fig:architecture}
\end{figure}

Figure~\ref{fig:architecture} summarizes the complete CORD framework,
including type-specific observation interfaces, shared structure--dynamics
representation learning, and target-aware downstream adaptation.

\subsection{Type-specific observation interfaces and shared backbone}
For observation $t$ in dataset $k$ and system type $d$, preprocessing constructs
local descriptors $X_t^{d,k}\in\mathbb R^{C_{d,k}\times W\times F}$,
global descriptors $G_t^{d,k}\in\mathbb R^{C_{d,k}\times F}$, and validity masks,
where $W$ is the number of local windows and $F$ the descriptor dimension.
The channel count $C_{d,k}$ may vary across datasets; padding is masked and
excluded from representation computation. Descriptors need not share physical
semantics across system types. Separate type-specific local and global stems
map them to a common latent width $d$. The instantiated descriptor definitions
and architecture dimensions are reported in Appendices~\ref{app:descriptors}
and~\ref{app:optimization}.

\begin{figure}[t]
\centering\includegraphics[width=.9\linewidth]{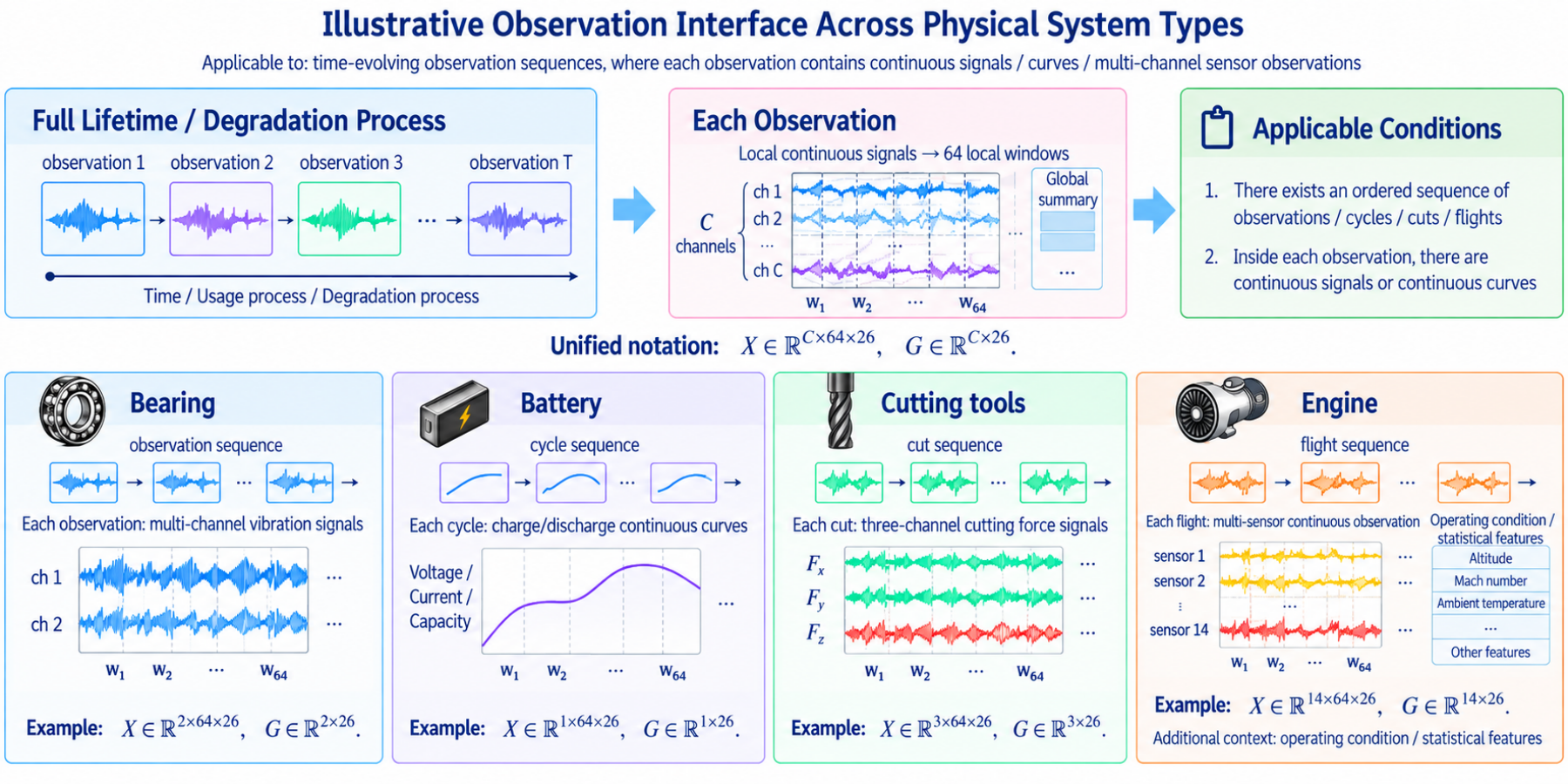}
\caption{Observation interfaces across physical system types. Bearing,
battery, cutting-tool, and engine observations retain different native signals
and channel semantics, while type-specific interfaces expose a common
local/global descriptor structure to the shared degradation backbone.}
\label{fig:observation-interface}
\end{figure}

Figure~\ref{fig:observation-interface} illustrates how different physical
system types retain their native observation sequences and channel semantics
while exposing a common local/global descriptor structure to the shared
backbone.
Validity-aware channel attention forms local and global tokens.
Shared Transformer blocks then process the tokens, while a type-specific
residual adapter follows each block:
\begin{equation}
 A_d(h)=W_{2,d}\operatorname{GELU}(W_{1,d}\operatorname{LN}(h)),
 \qquad h\leftarrow h+A_d(h).
\end{equation}
The global state and valid-local mean feed a shared observation projector.
Layer widths, adapter initialization, and readout dimensions are specified in
Appendix~\ref{app:optimization}.

\subsection{Structure--dynamics self-supervision}
CORD shapes the shared degradation representation at two complementary scales:
structure within the current observation and evolution across observation
histories.

\lead{Intra-Observation Structure Modeling (ISM).}
ISM masks a fraction $p_{\mathrm{mask}}$ of valid local tokens
($p_{\mathrm{mask}}=0.30$ in our experiments) and reconstructs observed
descriptor components. The reconstruction decoder is type-specific. If
$\Omega_{\mathrm{mask,obs}}$ indexes masked, observed entries, the objective is
\begin{equation}
 L_{\rm ISM}^{d,k}=\frac{1}{|\Omega_{\mathrm{mask,obs}}|}
 \sum_{(c,j,f)\in\Omega_{\mathrm{mask,obs}}}
 (\widehat X_{c,j,f}-X_{c,j,f})^2.
 \label{eq:ism}
\end{equation}
Only observed entries contribute, encouraging contextual descriptor structure
within an observation.

\lead{Inter-Observation Dynamics Modeling (IDM).}
IDM predicts the next observation embedding from an $H$-step history of the
same unit using a shared GRU and projection; the reported implementation uses
$H=6$:
\begin{equation}
 \widehat e_t=Q_{\psi}(\operatorname{GRU}_{\psi}(e_{t-H},\ldots,e_{t-1})),
 \qquad L_{\rm IDM}^{d,k}=\operatorname{MSE}(\widehat e_t,
 \operatorname{sg}(e_t)).
 \label{eq:idm}
\end{equation}
Gradients are stopped through the target embedding. ISM and IDM update the same encoding system at
different observation scales, coupling state structure and state evolution in
the shared encoder.

\subsection{Multi-domain pretraining and target-aware adaptation}
\lead{Coordinated source pretraining.}
Fixed initial, dataset-specific scales normalize both objectives. Type-specific
parameters receive both; shared parameters $\theta$ receive per-type gradients
\begin{equation}
 g_d=\rho_d\nabla_\theta\widetilde L_{\rm ISM}^{d,k}
       +\nabla_\theta\widetilde L_{\rm IDM}^{d,k},
 \label{eq:routing}
\end{equation}
Conflict-aware coordination aggregates shared gradients across system types
\citep{liu2021cagrad}; implementation details are provided in
Appendix~\ref{app:optimization}. One
selected source checkpoint initializes the three represented physical system
types.

\lead{Target-aware adaptation and evaluation regimes.}
The downstream prediction heads are target-aware to exploit the reusable
representation under each system's temporal structure. Bearings use a short
observation-history GRU, batteries use cycle history and trend, cutting tools use
hidden-state fusion and causal temporal modeling, and engines use flight history
and operating context. We evaluate three initialization regimes under the transfer
boundaries defined in Section~\ref{sec:problem}: \CordScratch starts from random
initialization, \CordSingle uses pretraining from the target physical system type
only, and \CordMulti uses joint pretraining across all represented system types.
All three use the same downstream architecture. During target adaptation, \Frozen
updates only the head, \PartialFT updates selected encoder components and the head,
and \FullFT updates all active encoder parameters and the head. For a
pretraining-excluded system type, a new type-specific input interface is randomly
initialized, while the shared encoder is initialized from source pretraining.
\section{Experimental Setup}
\subsection{Data and source-pretraining boundaries}\label{sec:protocol}
Source pretraining uses 19 datasets across three physical system types: seven
bearing, seven battery, and five cutting-tool datasets. Table~\ref{tab:setting}
lists the source pools, downstream boundaries, and held-out units. Level-I
target datasets are excluded in their entirety while other datasets of the
same type are used upstream; Level II excludes the complete turbofan-engine
type. Individual source-dataset citations are consolidated in
Appendix~\ref{app:protocol}.

\begin{table}[!htbp]
\centering
\caption{Source-pretraining pools and transfer boundaries. Level-I targets are
excluded upstream; Level II excludes the entire turbofan-engine system type.}
\label{tab:setting}
\begingroup\maintablefont
\begin{tabularx}{\linewidth}{@{}l>{\raggedright\arraybackslash}X>{\raggedright\arraybackslash}p{.40\linewidth}@{}}
\toprule
\rowcolor{cordHeader}
Level / type & Source-pretraining pool & Target boundary: adaptation $\rightarrow$ test \\
\midrule
I / Bearings & CWRU; FEMTO; Ferrara; IMS; KAIST; SEU; UNSW & XJTU-SY~\citep{wang2020hybrid}: Bearing2\_2--2\_5 $\rightarrow$ Bearing2\_1 \\
I / Batteries & HUST; Michigan; NASA; Oxford; KIT; SDU; XJTU & CALCE CS2~\citep{calceBatteryData}: CS2\_35--37 $\rightarrow$ CS2\_38 \\
I / Cutting tools & LUH; MATWI; Nonastreda; QIT-CEMC; HMoTP & PHM2010~\citep{phmsociety2010challenge}: C1,C4 $\rightarrow$ C6 \\
\addlinespace[2pt]
\rowcolor{cordTealLight}
II / Turbofan engines & None & N-CMAPSS~\citep{ariaschao2021ncmapss}: U2/U5/U10/U16/U18/U20 $\rightarrow$ U11 \\
\bottomrule
\end{tabularx}

\endgroup
\end{table}

\subsection{Training regimes, baselines, and reporting}
The same architecture is evaluated as \CordScratch, \CordSingle, or
\CordMulti. External baselines include MLP~\citep{rumelhart1986learning},
random forest~\citep{breiman2001random}, XGBoost~\citep{chen2016xgboost},
TCN~\citep{bai2018tcn}, PatchTST~\citep{nie2023patchtst},
iTransformer~\citep{liu2024itransformer}, and MOMENT~\citep{goswami2024moment}
with model-appropriate inputs and readouts.
We report mean $\pm$ SD over five downstream seeds at 10\%, 20\%, and 100\%
label budgets. Level-I runs use validation-selected checkpoints. For N-CMAPSS,
a fixed 400-update supervised adaptation budget is chosen by grouped
cross-validation on the six training engines and then applied to held-out U11
without validation-based early stopping.

\section{Results}

\subsection{Generalization under Pretraining-Included System Types}\label{sec:main}
Table~\ref{tab:reuse} compares CORD regimes on datasets never used in source
pretraining. \CordMulti improves over \CordSingle in all nine \Frozen
settings; the cutting-tool RMSE reductions are 30.6\%, 30.9\%, and 41.9\%.
Under \FullFT, \CordMulti improves over \CordSingle in eight of nine settings.
These results show that multi-domain pretraining improves reuse of the learned
representation, with consistent gains even when the pretrained encoder is kept
fixed.

\begin{table}[!htbp]
\centering
\caption{Level-I mean RMSE. \best{Best} and \second{second-best} values are
ranked within each system--budget column; complete metrics are in
Appendix~\ref{app:adaptation}.}
\label{tab:reuse}
\begingroup\maintablefont
% Mean RMSE from current_adaptation_{bearing,battery,milling}.tex.
% Best and second-best are ranked across all seven measured rows per system/budget.
\setlength{\tabcolsep}{2pt}
\begin{tabular}{llrrrrrrrrr}
\toprule
\rowcolor{cordHeader}
 & & \multicolumn{3}{c}{Bearings} & \multicolumn{3}{c}{Batteries} & \multicolumn{3}{c}{Cutting tools}\\
\cmidrule(lr){3-5}\cmidrule(lr){6-8}\cmidrule(lr){9-11}
\rowcolor{cordHeader}
Source & Adaptation & 10\% & 20\% & 100\% & 10\% & 20\% & 100\% & 10\% & 20\% & 100\%\\
\midrule
Scratch & --- & .1463 & .1400 & .1122 & .0711 & \second{.0680} & \best{.0597} & .1169 & .1007 & .0782\\
\midrule
Single-domain & Frozen & .1229 & .1395 & .1219 & .0739 & .0780 & .0657 & .1245 & .1131 & .1133\\
\rowcolor{cordRow}
Multi-domain & Frozen & \best{.1087} & \best{.1245} & .0991 & .0704 & \best{.0676} & .0636 & \second{.0864} & .0782 & .0658\\
Single-domain & Partial FT & .1211 & .1418 & .1034 & .0714 & .0818 & .0693 & .1194 & .1088 & .0920\\
\rowcolor{cordRow}
Multi-domain & Partial FT & .1175 & .1351 & .1063 & .0726 & .0732 & .0739 & \best{.0791} & \second{.0660} & \best{.0491}\\
Single-domain & Full FT & \second{.1121} & .1464 & \second{.0988} & \second{.0689} & .0735 & .0652 & .1082 & .0930 & .0800\\
\rowcolor{cordRow}
Multi-domain & Full FT & .1240 & \second{.1329} & \best{.0972} & \best{.0673} & .0683 & \second{.0632} & .0871 & \best{.0609} & \second{.0530}\\
\bottomrule
\end{tabular}

\endgroup
\end{table}

\begin{figure}[!htbp]
\centering\includegraphics[width=.88\linewidth,trim=0 10pt 0 6pt,clip]{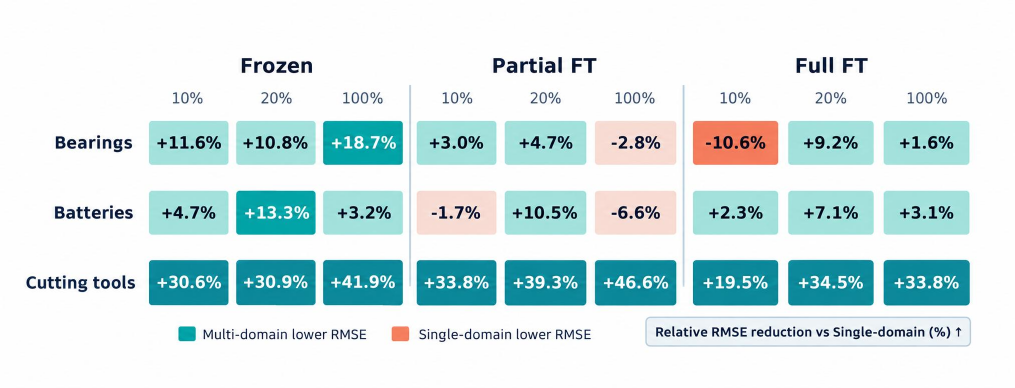}
\caption{Multi-domain pretraining across adaptation regimes. Cells show the
relative RMSE reduction of \CordMulti with respect to \CordSingle; the
in-figure key identifies which source regime has lower RMSE.}
\label{fig:within}
\end{figure}

Figure~\ref{fig:within} summarizes the relative effect of Multi-domain versus
Single-domain pretraining across \Frozen, \PartialFT, and \FullFT.
Multi-domain pretraining yields lower RMSE in all \Frozen settings, providing
the clearest evidence that the jointly learned representation is more reusable
without encoder updates. It retains an advantage in most \PartialFT and
\FullFT settings, although the gains are less uniform after fine-tuning.
Cutting tools show the largest and most consistent benefits across all three
adaptation strategies.

Table~\ref{tab:baseline} compares CORD with representative RUL prediction baselines. The pretrained CORD
rows use \FullFT as a common adaptation protocol, Scratch starts from random
initialization, and MOMENT is evaluated with a frozen encoder. Other
methods use their model-appropriate protocols; complete uncertainty and
metrics are in Appendix~\ref{app:baseline}.

\begin{table}[!htbp]
\centering
\caption{RUL prediction comparison: mean RMSE. CORD (Single-domain) and
CORD (Multi-domain) use \FullFT; Scratch trains from random initialization.
Each system--budget column ranks all displayed rows.}
\label{tab:baseline}
\begingroup\maintablefont
\setlength{\tabcolsep}{2pt}
\begin{tabular}{lrrrrrrrrr}
\toprule
\rowcolor{cordHeader}
& \multicolumn{3}{c}{Bearings} & \multicolumn{3}{c}{Batteries} & \multicolumn{3}{c}{Cutting tools}\\
\cmidrule(lr){2-4}\cmidrule(lr){5-7}\cmidrule(lr){8-10}
\rowcolor{cordHeader}
Model & 10\% & 20\% & 100\% & 10\% & 20\% & 100\% & 10\% & 20\% & 100\%\\
\midrule
CORD (Scratch) & .1463 & \second{.1400} & .1122 & .0711 & \best{.0680} & \best{.0597} & .1169 & .1007 & .0782\\
\rowcolor{cordRow}
CORD (Single-domain) & \best{.1121} & .1464 & \second{.0988} & .0689 & .0735 & .0652 & .1082 & \second{.0930} & .0800\\
\rowcolor{cordTealLight}
CORD (Multi-domain) & \second{.1240} & \best{.1329} & \best{.0972} & \best{.0673} & \second{.0683} & .0632 & \best{.0871} & \best{.0609} & \best{.0530}\\
\midrule
\rowcolor{cordRow}
MLP & .2185 & .2307 & .2160 & .1526 & .1408 & .1005 & \second{.0991} & .1085 & .1177\\
Random forest & .1840 & .1910 & .1853 & .1085 & .1110 & .1054 & .1117 & .1202 & .1230\\
\rowcolor{cordRow}
XGBoost & .1494 & .1824 & .1846 & .1036 & .1052 & .0992 & .1258 & .1204 & .1214\\
TCN & .2681 & .2930 & .2287 & .0807 & .0813 & .0788 & .1409 & .1266 & \second{.0722}\\
\rowcolor{cordRow}
PatchTST & .3019 & .3338 & .2663 & \second{.0676} & .0722 & .0792 & .2258 & .2103 & .1810\\
iTransformer & .3138 & .3157 & .3071 & .0904 & .0893 & .0813 & .2728 & .2582 & .1890\\
\rowcolor{cordRow}
MOMENT & .3070 & .3176 & .2711 & .0730 & .0695 & \second{.0620} & .2452 & .2431 & .2421\\
\bottomrule
\end{tabular}

\endgroup
\end{table}

A CORD regime achieves the lowest displayed RMSE in all nine system--budget
columns, with \CordMulti obtaining the best result in six of nine and outperforming
the representative external baselines across the table. The gains are
most pronounced in low-label settings, particularly for cutting tools. As target
supervision increases, \CordScratch becomes competitive on some targets, consistent
with pretraining providing the largest benefit when labeled target data are limited.

\subsection{Generalization under Pretraining-Excluded System Types}\label{sec:engine}
N-CMAPSS turbofan engines are entirely absent from source pretraining. Level II
uses a fixed 400-update training budget because labeled target data are more
limited, avoiding an additional per-seed validation split. The 400-update budget
is selected using grouped cross-validation on the six training engines only and
then fixed for evaluation on the held-out U11 engine.
Table~\ref{tab:engine-main} and Figure~\ref{fig:engine} compare Scratch and
source-pretrained initialization under the same 400-update budget. Source-pretrained
initialization reduces U11 RMSE from .1463 to .1374 at 10\% labels (6.1\%) and
from .1163 to .1122 at 20\% (3.4\%); MAE and $R^2$ show the same trend. At
100\% labels, Scratch achieves the lower RMSE (.0768 versus .0841). Overall, the
benefit of source pretraining is strongest in the low-label regime.

\begin{table}[!htbp]
\centering
\caption{Level-II N-CMAPSS U11 results under matched 400-update adaptation
(mean $\pm$ SD over five seeds).}
\label{tab:engine-main}
\begingroup\maintablefont\setlength{\tabcolsep}{2pt}
% Level-II summary: matched 400-update adaptation, five seeds.
\begin{tabular}{crrrrrr}
\toprule
\rowcolor{cordHeader}
& \multicolumn{2}{c}{RMSE $\downarrow$} & \multicolumn{2}{c}{MAE $\downarrow$} & \multicolumn{2}{c}{$R^2$ $\uparrow$}\\
\cmidrule(lr){2-3}\cmidrule(lr){4-5}\cmidrule(lr){6-7}
\rowcolor{cordHeader}
Labels & Scratch & \shortstack{Source-\\pretrained} & Scratch &
\shortstack{Source-\\pretrained} & Scratch & \shortstack{Source-\\pretrained}\\
\midrule
10\% & \second{.1463 $\pm$ .0134} & \best{.1374 $\pm$ .0117} & \second{.1151 $\pm$ .0103} & \best{.1098 $\pm$ .0091} & \second{.6518 $\pm$ .0611} & \best{.6932 $\pm$ .0510}\\
\rowcolor{cordRow}
20\% & \second{.1163 $\pm$ .0131} & \best{.1122 $\pm$ .0069} & \second{.0903 $\pm$ .0090} & \best{.0883 $\pm$ .0070} & \second{.7795 $\pm$ .0482} & \best{.7959 $\pm$ .0252}\\
100\% & \best{.0768 $\pm$ .0154} & \second{.0841 $\pm$ .0099} & \best{.0566 $\pm$ .0112} & \second{.0582 $\pm$ .0055} & \best{.9018 $\pm$ .0390} & \second{.8844 $\pm$ .0274}\\
\bottomrule
\end{tabular}

\endgroup
\end{table}

\subsection{Multi-domain pretraining improves the frozen representation}\label{sec:representation}
\begin{figure}[!t]
\centering
\begin{minipage}[t]{.48\linewidth}
\centering\includegraphics[width=.94\linewidth]{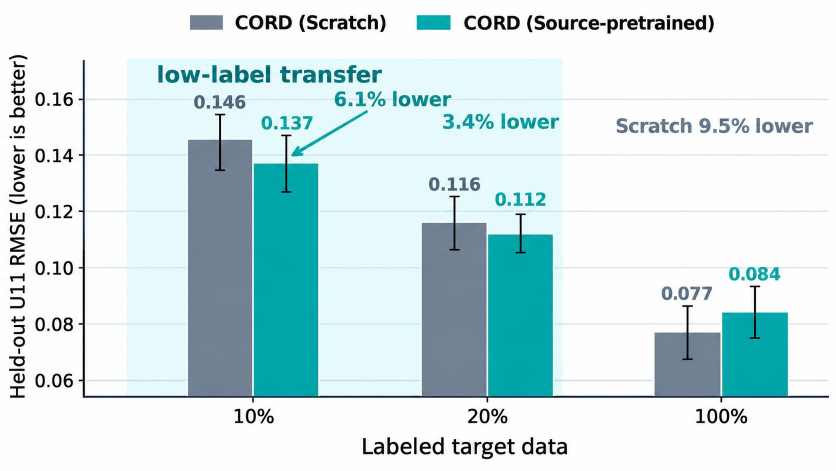}
\caption{Pretraining-excluded engine transfer. Held-out U11 RMSE under matched
400-update adaptation; whiskers show one sample SD.}
\label{fig:engine}
\end{minipage}\hfill
\begin{minipage}[t]{.48\linewidth}
\centering\includegraphics[width=\linewidth]{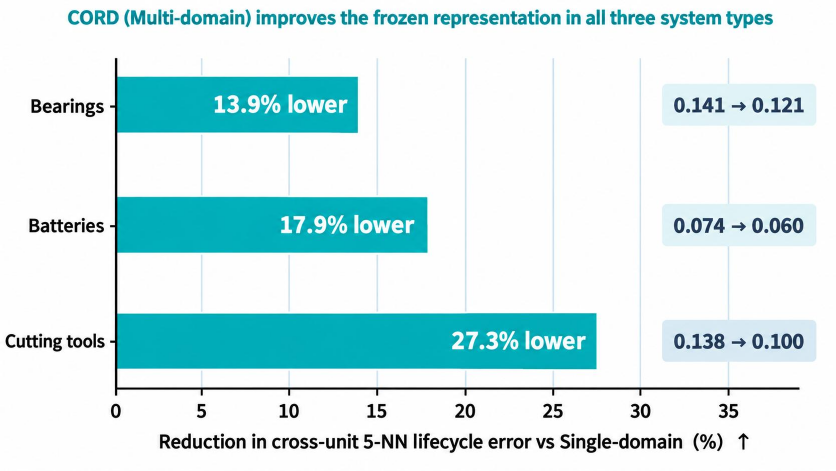}
\caption{Multi-domain pretraining improves frozen representation reuse. Bars
show relative reductions; callouts show absolute changes.}
\label{fig:representation}
\end{minipage}
\end{figure}

\begin{figure}[!t]
\centering\includegraphics[width=.88\linewidth]{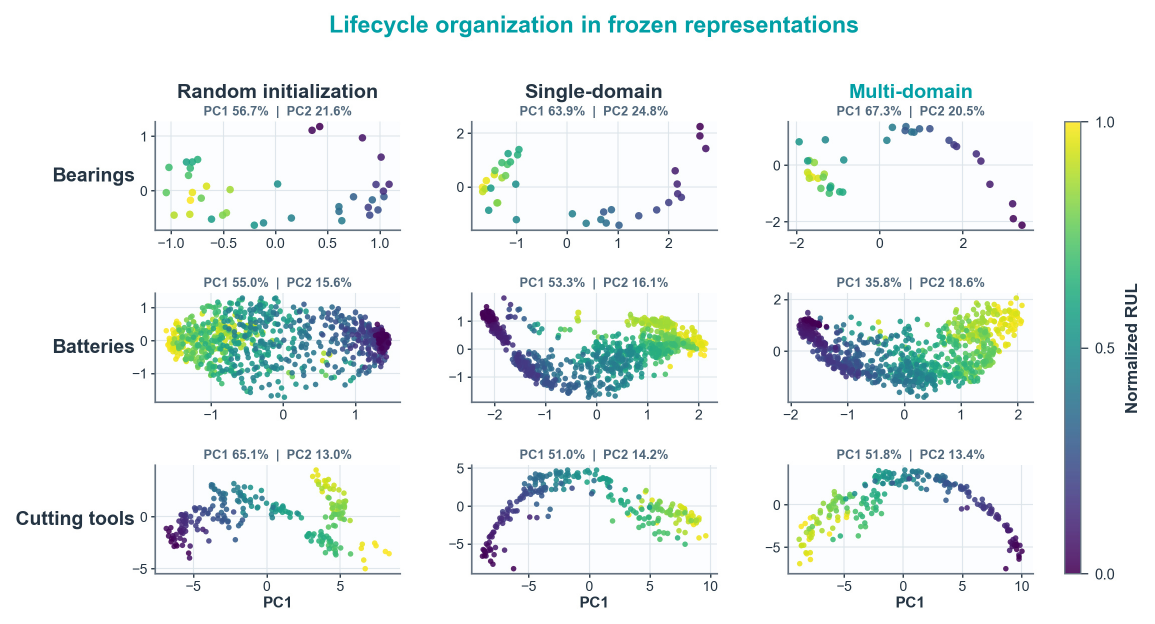}
\caption{Lifecycle organization in frozen representations. Points are colored
by normalized RUL. PCA is fitted independently within each encoder condition
and is used only to visualize within-panel lifecycle organization.}
\label{fig:pca}
\end{figure}

Before fitting the RUL heads, we evaluate whether the frozen representation
captures lifecycle structure consistently across units. Figure~\ref{fig:representation}
provides a quantitative measure using 5-NN lifecycle error, defined as the
normalized-RUL discrepancy between a held-out query and its five nearest
reference-unit states. \CordMulti reduces this error relative to \CordSingle by
13.9\% for bearings, 17.9\% for batteries, and 27.3\% for cutting tools, indicating
more lifecycle-consistent local neighborhoods. Figure~\ref{fig:pca} provides a complementary qualitative view of the same
representation space. Under multi-domain pretraining, samples within each system
type show a clearer and more continuous progression with normalized RUL, consistent
with the quantitative reductions in 5-NN lifecycle error.

\subsection{Ablations on Observation Interfaces and Pretraining Objectives}\label{sec:ablation}
\paragraph{Observation interface.}
Table~\ref{tab:ablation-main} compares the proposed \textbf{Structured
Descriptors} with \textbf{Raw-Resampled Inputs} under the same Scratch
protocol. Structured Descriptors achieve lower RMSE in all nine system--budget
settings, with Raw-Resampled Inputs yielding 2.16--3.00$\times$ larger errors.
This result shows that the performance gain is mainly attributed to the
structured observation interface rather than input dimensionality.

Figure~\ref{fig:observation-interface} illustrates how the same local/global
observation structure accommodates native bearing, battery, cutting-tool, and
engine measurements before they enter the shared representation backbone.

\begin{table}[!htbp]
\centering
\caption{Observation-interface ablation under the matched Scratch protocol.
Structured Descriptors are compared with parameter-free Raw-Resampled Inputs
using the same input dimensionality and downstream architecture. Entries
report mean RMSE over five downstream seeds; lower is better.}
\label{tab:ablation-main}
\begingroup
\maintablefont
\setlength{\tabcolsep}{2pt}
% Observation-interface ablation.
\begin{tabular}{lrrrrrrrrr}
\toprule
\rowcolor{cordHeader}
& \multicolumn{3}{c}{Bearings} & \multicolumn{3}{c}{Batteries} & \multicolumn{3}{c}{Cutting tools}\\
\cmidrule(lr){2-4}\cmidrule(lr){5-7}\cmidrule(lr){8-10}
\rowcolor{cordHeader}
Input construction & 10\% & 20\% & 100\% & 10\% & 20\% & 100\% & 10\% & 20\% & 100\%\\
\midrule
Structured Descriptors & \best{.1463} & \best{.1400} & \best{.1122} & \best{.0711} & \best{.0680} & \best{.0597} & \best{.1169} & \best{.1007} & \best{.0782}\\
\rowcolor{cordRow}
Raw-Resampled Inputs & \second{.3590} & \second{.3157} & \second{.3091} & \second{.2131} & \second{.1715} & \second{.1724} & \second{.2845} & \second{.2178} & \second{.1894}\\
\bottomrule
\end{tabular}

\endgroup
\end{table}

\paragraph{Inter-observation dynamics (IDM).}
Table~\ref{tab:idm-ablation-main} evaluates the contribution of IDM to transfer
on the pretraining-excluded turbofan-engine type using N-CMAPSS U11.

\begin{table}[!htbp]
\centering
\caption{Pretraining-objective ablation on N-CMAPSS U11. ISM-only and ISM+IDM
source-pretrained encoders use the same fixed 400-update adaptation protocol.
Entries are mean RMSE $\pm$ sample SD over five seeds; lower is better.}
\label{tab:idm-ablation-main}
\begingroup\maintablefont\setlength{\tabcolsep}{2pt}
\begin{tabular}{lccc}
\toprule
\rowcolor{cordHeader}
Pretraining objective & 10\% labels & 20\% labels & 100\% labels \\
\midrule
ISM only & \second{0.1725 $\pm$ 0.0156} & \second{0.1448 $\pm$ 0.0059} & \second{0.0944 $\pm$ 0.0189} \\
\rowcolor{cordRow}
ISM + IDM & \best{0.1374 $\pm$ 0.0117} & \best{0.1122 $\pm$ 0.0069} & \best{0.0841 $\pm$ 0.0099} \\
RMSE reduction & \textbf{20.3\%} & \textbf{22.5\%} & \textbf{10.9\%} \\
\bottomrule
\end{tabular}

\endgroup
\end{table}

Adding IDM reduces mean RMSE by 20.3\%, 22.5\%, and 10.9\% at 10\%, 20\%,
and 100\% labels, respectively. These results show that inter-observation
dynamics provide complementary information beyond within-observation structure,
with the largest gains under low-label cross-type adaptation. Complete MAE and
$R^2$ results are reported in Appendix~\ref{app:engine}.

\subsection{Efficiency and reusable deployment}\label{sec:efficiency}
CORD \FullFT predictors contain only 0.28--0.38M parameters across the three
Level-I systems. With one frozen shared encoder and three heads, deployment
uses 657,235 parameters versus 1,022,899 for three extracted copies, reducing
parameter storage by 35.75\%. These results show that representation reuse can
be achieved with a compact shared backbone. Detailed timing and memory results
are reported in Appendix~\ref{app:efficiency}.

\section{Discussion: What Makes Degradation Representations Reusable?}

\lead{Type-specific interfaces enable cross-system representation learning.}
CORD does not require physical system types to share sensors or variables.
Type-specific interfaces preserve system-dependent sensing semantics while
mapping observations into a shared degradation backbone. The
observation-interface ablation shows that the structured observation interface
consistently outperforms matched Raw-Resampled Inputs across the three
represented system types.

\lead{Structure and dynamics provide complementary pretraining signals.}
ISM models structural dependencies within individual observations, while IDM
models latent state evolution across observation histories. The Level II
ablation on the pretraining-excluded engine type shows that adding IDM reduces
prediction error across all label budgets, indicating that inter-observation
dynamics provide complementary information beyond within-observation
structure.

\lead{Multi-domain pretraining improves representation reuse across system boundaries.}
At Level I, \CordMulti outperforms \CordSingle in all nine \Frozen settings,
while cross-unit 5-NN lifecycle error also decreases for bearings, batteries,
and cutting tools. At Level II, source-pretrained initialization improves
adaptation to the pretraining-excluded engine type at 10\% and 20\% labels,
while its advantage diminishes at 100\%. These results show that multi-domain
pretraining supports representation reuse both within represented system types
and beyond the source-pretraining system boundary.

\section{Conclusion}

CORD separates heterogeneous observation modeling from shared degradation
representation learning. Type-specific interfaces preserve system-dependent
measurement semantics, while ISM and IDM capture within-observation structure
and inter-observation dynamics in a shared backbone. At Level I, multi-domain
pretraining improves frozen representation reuse on held-out bearing, battery,
and cutting-tool datasets. At Level II, source pretraining improves low-label
adaptation to a pretraining-excluded engine type. Overall, the results
demonstrate the feasibility of reusable degradation representation learning
across heterogeneous physical systems.
\label{maintextend}
\subsection*{AI use statement}
In this work, we used generative AI tools to assist with literature search,
research methodology and experiment design, code implementation and debugging,
result analysis and interpretation, figure preparation, and manuscript
organization, drafting, and language refinement. We did not use generative AI
tools to generate experimental data or to replace the execution and evaluation
of the reported experiments. We reviewed all AI-assisted work: literature
suggestions were checked against the original sources, generated code was
manually inspected and tested, numerical claims were verified against
experimental outputs, and AI-assisted text and figures were reviewed and
revised by the authors. We take responsibility for the final content of this
work, including text, claims, code, analyses, and artifacts produced with the
aid of generative AI.
\subsection*{Reproducibility statement}
The experiment packages retain preprocessing implementations, protocol files,
checkpoints, per-seed metrics, and prediction arrays.
Appendices~\ref{app:protocol} and~\ref{app:optimization} document the data
construction, architecture, optimization, and model-selection protocols, while
Appendices~\ref{app:adaptation}--\ref{app:efficiency} provide complete results
and computational details. The implementations, experiment configurations,
and retained experiment records are available in the
\href{https://github.com/HelpLee/CORD-PHM/}{GitHub repository}.
\subsection*{Ethics statement}
The intended application is research on equipment health monitoring. Prediction
errors can affect maintenance decisions; these experiments do not establish
safety for autonomous operational deployment. Any release of derived data or
model artifacts will comply with dataset licenses and redistribution permissions.
\bibliography{references}

@inproceedings{nie2023patchtst,
 title={A Time Series is Worth 64 Words: Long-term Forecasting with Transformers},
 author={Nie, Yuqi and Nguyen, Nam H. and Sinthong, Phanwadee and Kalagnanam, Jayant},
 booktitle={International Conference on Learning Representations}, year={2023},
 url={https://arxiv.org/abs/2211.14730}}

@article{wang2020hybrid,
 title={A Hybrid Prognostics Approach for Estimating Remaining Useful Life of Rolling Element Bearings},
 author={Wang, Biao and Lei, Yaguo and Li, Naipeng and Li, Ningbo},
 journal={IEEE Transactions on Reliability}, volume={69}, number={1},
 pages={401--412}, year={2020}, doi={10.1109/TR.2018.2882682},
 url={https://doi.org/10.1109/TR.2018.2882682}}

@misc{calceBatteryData,
 title={{CALCE} Battery Data: {CS2} Prismatic Cells},
 author={{Center for Advanced Life Cycle Engineering}},
 howpublished={University of Maryland battery data archive},
 year={n.d.},
 url={https://calce.umd.edu/data}}

@misc{phmsociety2010challenge,
 title={2010 {PHM} Society Conference Data Challenge},
 author={{PHM Society}}, year={2010},
 url={https://phmsociety.org/phm_competition/2010-phm-society-conference-data-challenge/}}

@article{rumelhart1986learning,
 title={Learning Representations by Back-Propagating Errors},
 author={Rumelhart, David E. and Hinton, Geoffrey E. and Williams, Ronald J.},
 journal={Nature}, volume={323}, number={6088}, pages={533--536}, year={1986},
 doi={10.1038/323533a0}, url={https://doi.org/10.1038/323533a0}}

@article{breiman2001random,
 title={Random Forests}, author={Breiman, Leo},
 journal={Machine Learning}, volume={45}, number={1}, pages={5--32}, year={2001},
 doi={10.1023/A:1010933404324}, url={https://doi.org/10.1023/A:1010933404324}}

@inproceedings{chen2016xgboost,
 title={{XGBoost}: A Scalable Tree Boosting System},
 author={Chen, Tianqi and Guestrin, Carlos},
 booktitle={Proceedings of the 22nd ACM SIGKDD International Conference on Knowledge Discovery and Data Mining},
 pages={785--794}, year={2016}, doi={10.1145/2939672.2939785},
 url={https://doi.org/10.1145/2939672.2939785}}

@misc{bai2018tcn,
 title={An Empirical Evaluation of Generic Convolutional and Recurrent Networks for Sequence Modeling},
 author={Bai, Shaojie and Kolter, J. Zico and Koltun, Vladlen},
 year={2018}, eprint={1803.01271}, archivePrefix={arXiv}, primaryClass={cs.LG},
 url={https://arxiv.org/abs/1803.01271}}

@inproceedings{liu2024itransformer,
 title={{iTransformer}: Inverted Transformers Are Effective for Time Series Forecasting},
 author={Liu, Yong and Hu, Tengge and Zhang, Haoran and Wu, Haixu and Wang, Shiyu and Ma, Lintao and Long, Mingsheng},
 booktitle={International Conference on Learning Representations}, year={2024},
 url={https://arxiv.org/abs/2310.06625}}

@inproceedings{he2026gtm,
 title={{GTM}: A General Time-series Model for Enhanced Representation Learning of Time-Series Data},
 author={He, Cheng and Huang, Xu and Jiang, Gangwei and Li, Zhaoyi and Lian, Defu and Xie, Hong and Chen, Enhong and Liang, Xijie and Zheng, Zengrong and Lee, Patrick P. C.},
 booktitle={International Conference on Learning Representations}, year={2026},
 url={https://proceedings.iclr.cc/paper_files/paper/2026/hash/0c6639f49f01a8578675303ce0030233-Abstract-Conference.html}}

@inproceedings{liu2021cagrad,
 title={Conflict-Averse Gradient Descent for Multi-task Learning},
 author={Liu, Bo and Liu, Xingchao and Jin, Xiaojie and Stone, Peter and Liu, Qiang},
 booktitle={Advances in Neural Information Processing Systems}, year={2021},
 url={https://proceedings.neurips.cc/paper/2021/hash/9d27fdf2477ffbff837d73ef7ae23db9-Abstract.html}}

@inproceedings{goswami2024moment,
 title={{MOMENT}: A Family of Open Time-series Foundation Models},
 author={Goswami, Mononito and Szafer, Konrad and Choudhry, Arjun and Cai, Yifu and Li, Shuo and Dubrawski, Artur},
 booktitle={Proceedings of the 41st International Conference on Machine Learning},
 volume={235}, pages={16115--16152}, year={2024},
 url={https://proceedings.mlr.press/v235/goswami24a.html}}

@inproceedings{woo2024moirai,
 title={Unified Training of Universal Time Series Forecasting Transformers},
 author={Woo, Gerald and Liu, Chenghao and Kumar, Akshat and Xiong, Caiming and Savarese, Silvio and Sahoo, Doyen},
 booktitle={Proceedings of the 41st International Conference on Machine Learning},
 volume={235}, pages={53140--53164}, year={2024}, url={https://proceedings.mlr.press/v235/woo24a.html}}

@inproceedings{shi2025timemoe,
 title={{Time-MoE}: Billion-Scale Time Series Foundation Models with Mixture of Experts},
 author={Shi, Xiaoming and Wang, Shiyu and Nie, Yuqi and Li, Dianqi and Ye, Zhou and Wen, Qingsong and Jin, Ming},
 booktitle={International Conference on Learning Representations}, year={2025},
 url={https://proceedings.iclr.cc/paper_files/paper/2025/hash/558d48c1f08675daa636e09bfe94a89e-Abstract-Conference.html}}

@inproceedings{rebuffi2017adapters,
 title={Learning Multiple Visual Domains with Residual Adapters},
 author={Rebuffi, Sylvestre-Alvise and Bilen, Hakan and Vedaldi, Andrea},
 booktitle={Advances in Neural Information Processing Systems}, year={2017},
 url={https://proceedings.neurips.cc/paper/2017/hash/e7b24b112a44fdd9ee93bdf998c6ca0e-Abstract.html}}

@inproceedings{houlsby2019adapter,
 title={Parameter-Efficient Transfer Learning for {NLP}},
 author={Houlsby, Neil and Giurgiu, Andrei and Jastrzebski, Stanislaw and Morrone, Bruna and de Laroussilhe, Quentin and Gesmundo, Andrea and Attariyan, Mona and Gelly, Sylvain},
 booktitle={Proceedings of the 36th International Conference on Machine Learning},
 volume={97}, pages={2790--2799}, year={2019},
 url={https://proceedings.mlr.press/v97/houlsby19a.html}}

@misc{theiler2026phm,
 title={Towards Unified and Data-Efficient Prognostics and Health Management with Tabular Foundation Models},
 author={Theiler, Raffael and Telyatnikov, Lev and von Krannichfeldt, Leandro and Fink, Olga},
 year={2026}, eprint={2606.05481}, archivePrefix={arXiv},
 url={https://arxiv.org/abs/2606.05481}}

@inproceedings{chen2026fedal,
 title={{FeDaL}: Federated Dataset Learning for General Time Series Foundation Models},
 author={Chen, Shengchao and Long, Guodong and Blumenstein, Michael and Jiang, Jing},
 booktitle={International Conference on Learning Representations}, year={2026},
 url={https://proceedings.iclr.cc/paper_files/paper/2026/hash/a0a53fefef4c2ad72d5ab79703ba70cb-Abstract-Conference.html}}

@inproceedings{huang2026formed,
 title={Repurposing Foundation Model for Generalizable Medical Time Series Classification},
 author={Huang, Nan and Wang, Haishuai and He, Zihuai and Zitnik, Marinka and Zhang, Xiang},
 booktitle={International Conference on Learning Representations}, year={2026},
 url={https://proceedings.iclr.cc/paper_files/paper/2026/file/8707924df5e207fa496f729f49069446-Paper-Conference.pdf}}

@inproceedings{fraikin2024trep,
 title={{T-Rep}: Representation Learning for Time Series using Time-Embeddings},
 author={Fraikin, Archibald and Bennetot, Adrien and Allassonniere, Stephanie},
 booktitle={International Conference on Learning Representations}, year={2024},
 url={https://proceedings.iclr.cc/paper_files/paper/2024/file/84b946c4c4162b29464fc7aa57e480e1-Paper-Conference.pdf}}

@article{dacosta2020rul,
 title={Remaining Useful Lifetime Prediction via Deep Domain Adaptation},
 author={da Costa, Paulo Roberto de Oliveira and Ak{\c c}ay, Alp and Zhang, Yingqian and Kaymak, Uzay},
 journal={Reliability Engineering \& System Safety}, volume={195}, pages={106682}, year={2020},
 doi={10.1016/j.ress.2019.106682}, url={https://doi.org/10.1016/j.ress.2019.106682}}

@article{wang2026heterogeneous,
 title={A Novel Two-Stage Heterogeneous Transfer Learning Framework for the Estimation of the Remaining Useful Life of Industrial Components},
 author={Wang, Bingsen and Baraldi, Piero and Shokry, Ahmed and Zio, Enrico},
 journal={Reliability Engineering \& System Safety}, year={2026},
 doi={10.1016/j.ress.2025.111968},
 url={https://doi.org/10.1016/j.ress.2025.111968}}

@article{ariaschao2021ncmapss,
 title={Aircraft Engine Run-to-Failure Dataset under Real Flight Conditions for Prognostics and Diagnostics},
 author={Arias Chao, Manuel and Kulkarni, Chetan and Goebel, Kai and Fink, Olga},
 journal={Data}, volume={6}, number={1}, pages={5}, year={2021},
 doi={10.3390/data6010005}, url={https://doi.org/10.3390/data6010005}}

@misc{cwruBearingData,
 author={{Case Western Reserve University Bearing Data Center}},
 title={{CWRU} Bearing Data Center},
 howpublished={Case School of Engineering}, year={n.d.},
 url={https://engineering.case.edu/bearingdatacenter/welcome}}

@article{smith2015cwru,
 author={Smith, Wade A. and Randall, Robert B.},
 title={Rolling Element Bearing Diagnostics Using the Case Western Reserve University Data: A Benchmark Study},
 journal={Mechanical Systems and Signal Processing}, year={2015}, volume={64--65}, pages={100--131},
 doi={10.1016/j.ymssp.2015.04.021}, url={https://doi.org/10.1016/j.ymssp.2015.04.021}}

@inproceedings{nectoux2012pronostia,
 author={Nectoux, Patrick and Gouriveau, Rafael and Medjaher, Kamal and Ramasso, Emmanuel and Chebel-Morello, Brigitte and Zerhouni, Noureddine and Varnier, Christophe},
 title={{PRONOSTIA}: An Experimental Platform for Bearings Accelerated Degradation Tests},
 booktitle={IEEE Conference on Prognostics and Health Management}, year={2012}, pages={1--8}}

@article{arpa2024ferrara,
 author={Arpa, Luca and Gabrielli, Alberto and Battarra, Mattia and Mucchi, Emiliano},
 title={University of {Ferrara} Run-to-Failure Vibration Dataset of Self-Aligning Double-Row Ball Bearings},
 journal={Data in Brief}, year={2024}, volume={55}, pages={110620},
 doi={10.1016/j.dib.2024.110620}, url={https://doi.org/10.1016/j.dib.2024.110620}}

@misc{qiu2007ims, author={Lee, J. and Qiu, H. and Yu, G. and Lin, J. and {Rexnord Technical Services}}, title={Bearing Data Set, {IMS}, University of Cincinnati}, year={2007}, howpublished={NASA Ames Prognostics Data Repository}, url={https://data.nasa.gov/dataset/ims-bearings}}

@article{jung2024kaist, author={Jung, Wonho and Yun, Sung-Hyun and Park, Yong-Hwa}, title={Vibration and Temperature Run-to-Failure Dataset of Ball Bearing for Prognostics}, journal={Data in Brief}, year={2024}, volume={54}, pages={110403}, doi={10.1016/j.dib.2024.110403}, url={https://doi.org/10.1016/j.dib.2024.110403}}

@article{shao2019seu, author={Shao, Siyu and McAleer, Stephen and Yan, Ruqiang and Baldi, Pierre}, title={Highly Accurate Machine Fault Diagnosis Using Deep Transfer Learning}, journal={IEEE Transactions on Industrial Informatics}, year={2019}, volume={15}, number={4}, pages={2446--2455}}

@article{zhang2022unsw, author={Zhang, Hengcheng and Borghesani, Pietro and Randall, Robert B. and Peng, Zhongxiao}, title={A Benchmark of Measurement Approaches to Track the Natural Evolution of Spall Severity in Rolling Element Bearings}, journal={Mechanical Systems and Signal Processing}, year={2022}, volume={166}, pages={108466}}

@article{ma2022hust, author={Ma, Guijun and Xu, Songpei and Jiang, Benben and others}, title={Real-Time Personalized Health Status Prediction of Lithium-Ion Batteries Using Deep Transfer Learning}, journal={Energy \& Environmental Science}, year={2022}, volume={15}, pages={4083--4094}}

@article{weng2021michigan, author={Weng, Andrew and Mohtat, Peyman and Attia, Peter M. and others}, title={Predicting the Impact of Formation Protocols on Battery Lifetime Immediately after Manufacturing}, journal={Joule}, year={2021}, volume={5}, number={11}, pages={2971--2992}}

@misc{saha2007nasa, author={Saha, Bhaskar and Goebel, Kai}, title={Battery Data Set}, year={2007}, howpublished={NASA Ames Prognostics Data Repository}}

@misc{howey2017oxford, author={Howey, David and Birkl, Christoph}, title={Oxford Battery Degradation Dataset 1}, year={2017}, howpublished={University of Oxford Research Archive}}

@article{luh2024kit, author={Luh, Matthias and Blank, Thomas}, title={Comprehensive Battery Aging Dataset: Capacity and Impedance Fade Measurements of a Lithium-Ion {NMC/C-SiO} Cell}, journal={Scientific Data}, year={2024}, volume={11}, pages={1004}, doi={10.1038/s41597-024-03831-x}, url={https://doi.org/10.1038/s41597-024-03831-x}}

@article{wang2025sdu, author={Wang, Shuquan and Gao, Feng and Tian, Hao}, title={Deep Sorting of Reused Batteries for Enabling Long-Term Consistency Grouping with Unknown Prior Conditions}, journal={Cell Reports Physical Science}, year={2025}, volume={6}, number={7}, pages={102657}}

@article{wang2024xjtubattery, author={Wang, Fujin and Zhai, Z. and Zhao, Z. and others}, title={Physics-Informed Neural Network for Lithium-Ion Battery Degradation Stable Modeling and Prognosis}, journal={Nature Communications}, year={2024}, volume={15}, pages={4332}}

@article{denkena2023luh, author={Denkena, Berend and Klemme, Heinrich and Stiehl, Tobias H.}, title={Multivariate Time Series Data of Milling Processes with Varying Tool Wear and Machine Tools}, journal={Data in Brief}, year={2023}, volume={50}, pages={109574}, doi={10.1016/j.dib.2023.109574}, url={https://doi.org/10.1016/j.dib.2023.109574}}

@inproceedings{depauw2023matwi, author={De Pauw, Lars and Jacobs, Tom and Goedem{\'e}, Toon}, title={{MATWI}: A Multimodal Automatic Tool Wear Inspection Dataset and Baseline Algorithms}, booktitle={International Conference on Computer Vision Systems}, year={2023}, pages={255--269}, doi={10.1007/978-3-031-44137-0_22}, url={https://doi.org/10.1007/978-3-031-44137-0_22}}

@article{truchan2025nonastreda, author={Truchan, Hubert and Ahmadi, Zahra}, title={Nonastreda Multimodal Dataset for Efficient Tool Wear State Monitoring}, journal={Data in Brief}, year={2025}, volume={62}, pages={111905}}

@article{li2025qit, author={Li, Na and Wang, Xiao and Wang, Wanzhen and Xin, Miaomiao and Yuan, Dongfeng and Zhang, Mingqiang}, title={A Multi-Feature Dataset of Coated End Milling Cutter Tool Wear Whole Life Cycle}, journal={Scientific Data}, year={2025}, volume={12}, pages={16}}

@article{wang2024hmotp, author={Wang, Runqiong and Song, Qinghua and Peng, Y. and others}, title={Toward Digital Twins for High-Performance Manufacturing: Tool Wear Monitoring in High-Speed Milling of Thin-Walled Parts Using Domain Knowledge}, journal={Robotics and Computer-Integrated Manufacturing}, year={2024}, volume={88}, pages={102723}}

@inproceedings{das2024timesfm,
  title={A Decoder-Only Foundation Model for Time-Series Forecasting},
  author={Das, Abhimanyu and Kong, Weihao and Sen, Rajat and Zhou, Yichen},
  booktitle={Proceedings of the 41st International Conference on Machine Learning},
  volume={235}, pages={10148--10167}, year={2024}, publisher={PMLR},
  url={https://proceedings.mlr.press/v235/das24c.html}}

@article{ansari2024chronos,
  title={{Chronos}: Learning the Language of Time Series},
  author={Ansari, Abdul Fatir and Stella, Lorenzo and Turkmen, Caner and Zhang, Xiyuan and Mercado, Pedro and Shen, Huibin and Shchur, Oleksandr and Rangapuram, Syama Sundar and Pineda Arango, Sebastian and Kapoor, Shubham and Zschiegner, Jasper and Maddix, Danielle C. and Wang, Hao and Mahoney, Michael W. and Torkkola, Kari and Wilson, Andrew Gordon and Bohlke-Schneider, Michael and Wang, Yuyang},
  journal={Transactions on Machine Learning Research}, year={2024},
  url={https://arxiv.org/abs/2403.07815}}

@inproceedings{liu2024timer,
  title={Timer: Generative Pre-Trained Transformers Are Large Time Series Models},
  author={Liu, Yong and Zhang, Haoran and Li, Chenyu and Huang, Xiangdong and Wang, Jianmin and Long, Mingsheng},
  booktitle={Proceedings of the 41st International Conference on Machine Learning},
  volume={235}, pages={32369--32399}, year={2024}, publisher={PMLR},
  url={https://proceedings.mlr.press/v235/liu24cb.html}}

@inproceedings{dong2024timesiam,
  title={{TimeSiam}: A Pre-Training Framework for Siamese Time-Series Modeling},
  author={Dong, Jiaxiang and Wu, Haixu and Wang, Yuxuan and Qiu, Yun-Zhong and Zhang, Li and Wang, Jianmin and Long, Mingsheng},
  booktitle={Proceedings of the 41st International Conference on Machine Learning},
  volume={235}, pages={11412--11436}, year={2024}, publisher={PMLR}}
\bibliographystyle{iclr2027_conference}
\clearpage
\appendix
% Start the appendix on a fresh page.  Do not vertically stretch sparse
% appendix pages: doing so creates conspicuous gaps around displays and tables.
\raggedbottom
% longtable otherwise restricts its caption to a 4-inch centred block.  Match
% ordinary appendix tables by giving every longtable caption the full text
% width; LaTeX then centres short captions and wraps long ones consistently.
\setlength{\LTcapwidth}{\textwidth}
% Appendix table captions use one visual rule regardless of caption length or
% whether the table is implemented with table or longtable.
\captionsetup[table]{width=\textwidth,justification=raggedright,
  singlelinecheck=false,skip=2pt}
\setlength{\abovedisplayskip}{7pt plus 2pt minus 2pt}
\setlength{\belowdisplayskip}{7pt plus 2pt minus 2pt}
\setlength{\abovedisplayshortskip}{4pt plus 2pt minus 1pt}
\setlength{\belowdisplayshortskip}{5pt plus 2pt minus 1pt}
\setlength{\intextsep}{7pt plus 2pt minus 2pt}
\setlength{\textfloatsep}{9pt plus 2pt minus 2pt}
\setlength{\floatsep}{7pt plus 2pt minus 2pt}
\setlength{\abovecaptionskip}{3pt}
\setlength{\belowcaptionskip}{2pt}
\section{Data Boundaries and Observation Construction}\label{app:protocol}
Source pretraining uses 19 datasets across the three represented physical system types.
The downstream datasets are excluded in their entirety from source pretraining.
Historical three-channel cutting-tool results are excluded from this study;
PHM2010 uses its seven recorded channels. Observation descriptors use 64 local
slots and 26 feature dimensions with validity masks. Target construction and
dataset-specific preprocessing are recorded in the released manifests.
Main-text Table~\ref{tab:setting} reports the source pools, transfer boundaries,
and downstream units in one place.

\begin{table}[H]
\centering\small
\caption{Source-pretraining and downstream target datasets with their direct
data-source citations. References are consolidated here to keep the main-text
data description readable.}
\label{tab:source-citations}
\begin{tabularx}{\linewidth}{@{}l l >{\raggedright\arraybackslash}X@{}}
\toprule
\rowcolor{cordHeader}
System type & Dataset & Reference\\
\midrule
Bearings & CWRU & \citep{cwruBearingData,smith2015cwru}\\
& FEMTO / PRONOSTIA & \citep{nectoux2012pronostia}\\
& Ferrara & \citep{arpa2024ferrara}\\
& IMS & \citep{qiu2007ims}\\
& KAIST & \citep{jung2024kaist}\\
& SEU & \citep{shao2019seu}\\
& UNSW & \citep{zhang2022unsw}\\
\midrule
Batteries & HUST & \citep{ma2022hust}\\
& Michigan & \citep{weng2021michigan}\\
& NASA & \citep{saha2007nasa}\\
& Oxford & \citep{howey2017oxford}\\
& KIT NMC/C-SiO & \citep{luh2024kit}\\
& SDU & \citep{wang2025sdu}\\
& XJTU & \citep{wang2024xjtubattery}\\
\midrule
Cutting tools & LUH & \citep{denkena2023luh}\\
& MATWI & \citep{depauw2023matwi}\\
& Nonastreda & \citep{truchan2025nonastreda}\\
& QIT-CEMC & \citep{li2025qit}\\
& HMoTP & \citep{wang2024hmotp}\\
\midrule
Downstream targets & XJTU-SY bearings & \citep{wang2020hybrid}\\
& CALCE CS2 batteries & \citep{calceBatteryData}\\
& PHM2010 cutting tools & \citep{phmsociety2010challenge}\\
& N-CMAPSS turbofan engines & \citep{ariaschao2021ncmapss}\\
\bottomrule
\end{tabularx}

\end{table}

\subsection{Structured Descriptor Definitions}\label{app:descriptors}
For the reported implementation, $F=26$; Table~\ref{tab:descriptor-definitions}
records the exact feature order used by the three represented system-type
interfaces. Identical names in the bearing and
cutting-tool columns denote the same statistic applied to different native sensor
channels; they do not impose shared channel semantics across system types.
\begingroup
\small
\setlength{\tabcolsep}{3pt}
\renewcommand{\arraystretch}{1.03}
\begin{longtable}{@{}r >{\raggedright\arraybackslash}p{0.235\linewidth} >{\raggedright\arraybackslash}p{0.34\linewidth} >{\raggedright\arraybackslash}p{0.235\linewidth}@{}}
\caption{Ordered 26-dimensional Structured Descriptors exposed by each source-system observation interface. Bearing and cutting-tool descriptors are computed independently for every valid native sensor channel and local signal window. Battery descriptors are computed over valid local discharge windows; unavailable temperature entries are masked rather than imputed as observations.}\label{tab:descriptor-definitions}\\
\toprule
\rowcolor{cordHeader}
Index & Bearings & Batteries & Cutting tools \\
\midrule
\endfirsthead
\multicolumn{4}{l}{\small\itshape Table~\thetable\ continued.}\\
\toprule
\rowcolor{cordHeader}
Index & Bearings & Batteries & Cutting tools \\
\midrule
\endhead
\midrule
\multicolumn{4}{r}{\small\itshape Continued on next page.}\\
\endfoot
\bottomrule
\endlastfoot
1  & Mean & Mean voltage & Mean \\
2  & Mean absolute value & Voltage standard deviation & Mean absolute value \\
3  & Standard deviation & Minimum voltage & Standard deviation \\
4  & Variance & Maximum voltage & Variance \\
5  & Root mean square & Voltage peak-to-peak range & Root mean square \\
6  & Mean-square energy & Voltage skewness & Mean-square energy \\
7  & Absolute peak amplitude & Voltage kurtosis & Absolute peak amplitude \\
8  & Peak-to-peak range & Voltage slope & Peak-to-peak range \\
9  & Minimum & Mean absolute voltage derivative & Minimum \\
10 & Maximum & Mean absolute voltage curvature & Maximum \\
11 & Skewness & Mean C-rate & Skewness \\
12 & Kurtosis & C-rate standard deviation & Kurtosis \\
13 & Crest factor & Minimum C-rate & Crest factor \\
14 & Shape factor & Maximum C-rate & Shape factor \\
15 & Impulse factor & Mean absolute C-rate & Impulse factor \\
16 & Clearance factor & C-rate slope & Clearance factor \\
17 & Root amplitude & Mean temperature & Root amplitude \\
18 & Zero-crossing rate & Temperature standard deviation & Zero-crossing rate \\
19 & Spectral centroid & Minimum temperature & Spectral centroid \\
20 & Spectral bandwidth & Maximum temperature & Spectral bandwidth \\
21 & Spectral entropy & Temperature change & Spectral entropy \\
22 & Dominant frequency & Temperature slope & Dominant frequency \\
23 & Low-band normalized energy & Local duration & Low-band normalized energy \\
24 & Mid-band normalized energy & Local capacity & Mid-band normalized energy \\
25 & High-band normalized energy & Local energy & High-band normalized energy \\
26 & High/low-band energy ratio & Mean $\mathrm{d}V/\mathrm{d}Q$ & High/low-band energy ratio \\
\end{longtable}
\endgroup

\section{Architecture, Optimization, and Model Selection}\label{app:optimization}
The reported implementation uses $W=64$, $F=26$, $d=96$, $h=4$,
$d_{\mathrm{ff}}=192$, and $r=24$. Type-specific local and global stems map
the descriptors to width $d$. Validity-aware channel attention yields $W$
local tokens plus one global token. Two shared pre-normalized Transformer
blocks use $h$ attention heads and feed-forward width $d_{\mathrm{ff}}$; a
residual adapter after each block has bottleneck width $r$ and a
zero-initialized final map. The global state and valid-local mean form a
$2d$-dimensional concatenation, which is projected back to $d$ to produce the
observation representation. Cutting-tool prediction may also use the
$2d$-dimensional final-normalized token readout.

The selected CORD checkpoint uses a bearing-gradient directional projection following
CAGrad. For dataset $k$ in system type $d$, initial source-training batches fix scales
$s_{\rm ISM}^{d,k}$ and $s_{\rm IDM}^{d,k}$:
\begin{equation}
 \widetilde L_{\rm ISM}^{d,k}=L_{\rm ISM}^{d,k}/(2s_{\rm ISM}^{d,k}),
 \qquad \widetilde L_{\rm IDM}^{d,k}=L_{\rm IDM}^{d,k}/(2s_{\rm IDM}^{d,k}).
\end{equation}
System-type-specific stems, adapters, channel embeddings, and decoders receive both
normalized components. For shared parameters, Eq.~\ref{eq:routing} uses fixed
ISM coefficients $(0.3,0.3,1)$ for bearings, batteries, and cutting tools;
IDM is not multiplied by $1-\rho_d$. CAGrad uses $\alpha=0.4$ and its
$\mathrm{rescale}=1$ convention. Given its output $v$ and bearing gradient
$g_b$, the implementation then applies
\begin{equation}
 v'=v+\frac{[\|g_b\|^2/D-g_b^\top v]_+}
                  {\max(\|g_b\|^2,10^{-20})}g_b,\qquad D=3.
 \label{eq:floor}
\end{equation}
This minimum Euclidean correction enforces the specified bearing-gradient
directional floor before clipping and AdamW. Single-domain training uses the
corresponding interface, scales, and coefficient without multi-domain gradient
aggregation.

Each epoch has 20 joint updates, each with batch 32 per active system type and
microbatch eight. Source datasets rotate within system types. AdamW uses learning rate
and weight decay $10^{-4}$, with a 2,000-epoch ceiling, patience 30, and minimum
validation improvement $10^{-4}$. Source selection averages dataset loss
ratios within system types, then averages system types. The validation total is reconstruction
plus $0.2$ times dynamics relative to its initial value; it differs from the
separately normalized training objective. One joint checkpoint is selected for
all represented system types.

The target-aware RUL readouts use nonlinear temporal heads. Bearings use
up to six observations and a GRU; batteries use a 20-cycle history/trend head;
cutting tools use up to 20 observations, hidden-state fusion, a causal TCN, and a
GRU. Engines use seven-flight histories and operating context. Represented-type
fine-tuning uses encoder learning rate $3\times10^{-4}$ and head rate
$10^{-3}$; \CordScratch uses $10^{-3}$. \PartialFT updates only
designated final encoder components.

\section{Generalization under Pretraining-Included System Types: Complete Results}\label{app:adaptation}
Tables report mean $\pm$ SD. \best{Red bold} marks
the lowest displayed mean and \second{blue bold underlined} the second-lowest
within each physical system type and label budget among the listed adaptation procedures.
\subsection{Bearing}
\begin{table}[H]
\centering\small
\caption{Bearing adaptation on XJTU-SY: RMSE, MAE, and $R^2$ at all three label budgets (five seeds).}
\label{tab:appendix-adaptation-bearing}
\begin{tabular}{lllll}
\toprule
\rowcolor{cordHeader}
Labels & Treatment & RMSE & MAE & $R^2$ \\
\midrule
10\% & CORD (Scratch) & $0.1463 \pm 0.0393$ & $0.1172 \pm 0.0292$ & $0.7428 \pm 0.1406$ \\
\rowcolor{cordRow}
10\% & CORD (Single-domain) Frozen & $0.1229 \pm 0.0176$ & $0.0984 \pm 0.0207$ & $0.8256 \pm 0.0519$ \\
10\% & CORD (Multi-domain) Frozen & $\best{0.1087 \pm 0.0153}$ & $\second{0.0859 \pm 0.0114}$ & $\best{0.8635 \pm 0.0379}$ \\
\rowcolor{cordRow}
10\% & CORD (Single-domain) Partial FT & $0.1211 \pm 0.0318$ & $0.0944 \pm 0.0261$ & $0.8242 \pm 0.0959$ \\
10\% & CORD (Multi-domain) Partial FT & $0.1175 \pm 0.0190$ & $0.0977 \pm 0.0155$ & $0.8397 \pm 0.0523$ \\
\rowcolor{cordRow}
10\% & CORD (Single-domain) Full FT & $\second{0.1121 \pm 0.0224}$ & $\best{0.0833 \pm 0.0142}$ & $\second{0.8526 \pm 0.0623}$ \\
10\% & CORD (Multi-domain) Full FT & $0.1240 \pm 0.0289$ & $0.0979 \pm 0.0239$ & $0.8177 \pm 0.0859$ \\
\midrule
\rowcolor{cordRow}
20\% & CORD (Scratch) & $0.1400 \pm 0.0145$ & $0.1081 \pm 0.0112$ & $0.7752 \pm 0.0453$ \\
20\% & CORD (Single-domain) Frozen & $0.1395 \pm 0.0185$ & $0.1141 \pm 0.0086$ & $0.7758 \pm 0.0581$ \\
\rowcolor{cordRow}
20\% & CORD (Multi-domain) Frozen & $\best{0.1245 \pm 0.0097}$ & $\best{0.0994 \pm 0.0095}$ & $\best{0.8229 \pm 0.0283}$ \\
20\% & CORD (Single-domain) Partial FT & $0.1418 \pm 0.0252$ & $0.1127 \pm 0.0210$ & $0.7656 \pm 0.0879$ \\
\rowcolor{cordRow}
20\% & CORD (Multi-domain) Partial FT & $0.1351 \pm 0.0171$ & $0.1108 \pm 0.0120$ & $0.7899 \pm 0.0516$ \\
20\% & CORD (Single-domain) Full FT & $0.1464 \pm 0.0142$ & $0.1100 \pm 0.0132$ & $0.7546 \pm 0.0474$ \\
\rowcolor{cordRow}
20\% & CORD (Multi-domain) Full FT & $\second{0.1329 \pm 0.0214}$ & $\second{0.1044 \pm 0.0142}$ & $\second{0.7951 \pm 0.0697}$ \\
\midrule
100\% & CORD (Scratch) & $0.1122 \pm 0.0233$ & $0.0854 \pm 0.0143$ & $0.8519 \pm 0.0637$ \\
\rowcolor{cordRow}
100\% & CORD (Single-domain) Frozen & $0.1219 \pm 0.0107$ & $0.0940 \pm 0.0086$ & $0.8301 \pm 0.0308$ \\
100\% & CORD (Multi-domain) Frozen & $0.0991 \pm 0.0166$ & $0.0805 \pm 0.0121$ & $0.8857 \pm 0.0371$ \\
\rowcolor{cordRow}
100\% & CORD (Single-domain) Partial FT & $0.1034 \pm 0.0182$ & $0.0802 \pm 0.0131$ & $0.8753 \pm 0.0455$ \\
100\% & CORD (Multi-domain) Partial FT & $0.1063 \pm 0.0190$ & $0.0842 \pm 0.0137$ & $0.8682 \pm 0.0488$ \\
\rowcolor{cordRow}
100\% & CORD (Single-domain) Full FT & $\second{0.0988 \pm 0.0184}$ & $\best{0.0718 \pm 0.0162}$ & $\second{0.8859 \pm 0.0436}$ \\
100\% & CORD (Multi-domain) Full FT & $\best{0.0972 \pm 0.0071}$ & $\second{0.0744 \pm 0.0076}$ & $\best{0.8921 \pm 0.0160}$ \\
\bottomrule
\end{tabular}

\end{table}
\subsection{Battery}
{\small
\setlength\LTleft{\fill}
\setlength\LTright{\fill}
\begin{longtable}{lllll}
\caption{Battery adaptation on CALCE CS2: RMSE, MAE, and $R^2$ at all three label budgets (five seeds).}
\label{tab:appendix-adaptation-battery}\\
\toprule
\rowcolor{cordHeader}
Labels & Treatment & RMSE & MAE & $R^2$ \\
\midrule
\endfirsthead
\toprule
\rowcolor{cordHeader}
Labels & Treatment & RMSE & MAE & $R^2$ \\
\midrule
\endhead
\midrule
\multicolumn{5}{r}{\footnotesize Continued on next page}\\
\endfoot
\bottomrule
\endlastfoot
10\% & CORD (Scratch) & $0.0711 \pm 0.0091$ & $0.0516 \pm 0.0065$ & $0.9384 \pm 0.0146$ \\
\rowcolor{cordRow}
10\% & CORD (Single-domain) Frozen & $0.0739 \pm 0.0059$ & $0.0547 \pm 0.0042$ & $0.9339 \pm 0.0106$ \\
10\% & CORD (Multi-domain) Frozen & $0.0704 \pm 0.0028$ & $0.0522 \pm 0.0019$ & $0.9402 \pm 0.0048$ \\
\rowcolor{cordRow}
10\% & CORD (Single-domain) Partial FT & $0.0714 \pm 0.0030$ & $0.0522 \pm 0.0019$ & $0.9385 \pm 0.0051$ \\
10\% & CORD (Multi-domain) Partial FT & $0.0726 \pm 0.0040$ & $0.0529 \pm 0.0031$ & $0.9364 \pm 0.0070$ \\
\rowcolor{cordRow}
10\% & CORD (Single-domain) Full FT & $\second{0.0689 \pm 0.0048}$ & $\second{0.0492 \pm 0.0035}$ & $\second{0.9426 \pm 0.0078}$ \\
10\% & CORD (Multi-domain) Full FT & $\best{0.0673 \pm 0.0064}$ & $\best{0.0487 \pm 0.0044}$ & $\best{0.9450 \pm 0.0105}$ \\
\midrule
\rowcolor{cordRow}
20\% & CORD (Scratch) & $\second{0.0680 \pm 0.0060}$ & $0.0497 \pm 0.0056$ & $\second{0.9439 \pm 0.0101}$ \\
20\% & CORD (Single-domain) Frozen & $0.0780 \pm 0.0042$ & $0.0561 \pm 0.0038$ & $0.9265 \pm 0.0080$ \\
\rowcolor{cordRow}
20\% & CORD (Multi-domain) Frozen & $\best{0.0676 \pm 0.0012}$ & $\second{0.0494 \pm 0.0009}$ & $\best{0.9450 \pm 0.0019}$ \\
20\% & CORD (Single-domain) Partial FT & $0.0818 \pm 0.0057$ & $0.0594 \pm 0.0038$ & $0.9191 \pm 0.0109$ \\
\rowcolor{cordRow}
20\% & CORD (Multi-domain) Partial FT & $0.0732 \pm 0.0052$ & $0.0524 \pm 0.0040$ & $0.9351 \pm 0.0092$ \\
20\% & CORD (Single-domain) Full FT & $0.0735 \pm 0.0050$ & $0.0533 \pm 0.0044$ & $0.9346 \pm 0.0089$ \\
\rowcolor{cordRow}
20\% & CORD (Multi-domain) Full FT & $0.0683 \pm 0.0043$ & $\best{0.0484 \pm 0.0035}$ & $0.9436 \pm 0.0072$ \\
\midrule
100\% & CORD (Scratch) & $\best{0.0597 \pm 0.0085}$ & $\best{0.0435 \pm 0.0073}$ & $\best{0.9564 \pm 0.0125}$ \\
\rowcolor{cordRow}
100\% & CORD (Single-domain) Frozen & $0.0657 \pm 0.0070$ & $0.0480 \pm 0.0055$ & $0.9475 \pm 0.0115$ \\
100\% & CORD (Multi-domain) Frozen & $0.0636 \pm 0.0018$ & $\second{0.0454 \pm 0.0026}$ & $0.9512 \pm 0.0028$ \\
\rowcolor{cordRow}
100\% & CORD (Single-domain) Partial FT & $0.0693 \pm 0.0013$ & $0.0504 \pm 0.0005$ & $0.9421 \pm 0.0022$ \\
100\% & CORD (Multi-domain) Partial FT & $0.0739 \pm 0.0085$ & $0.0527 \pm 0.0040$ & $0.9334 \pm 0.0159$ \\
\rowcolor{cordRow}
100\% & CORD (Single-domain) Full FT & $0.0652 \pm 0.0064$ & $0.0464 \pm 0.0054$ & $0.9483 \pm 0.0103$ \\
100\% & CORD (Multi-domain) Full FT & $\second{0.0632 \pm 0.0056}$ & $0.0458 \pm 0.0050$ & $\second{0.9515 \pm 0.0086}$ \\
\end{longtable}

}
\subsection{Cutting tools}
\begin{table}[H]
\centering\small
\caption{Cutting-tool adaptation on PHM2010: RMSE, MAE, and $R^2$ at all three label budgets (five seeds).}
\label{tab:appendix-adaptation-milling}
\begin{tabular}{lllll}
\toprule
\rowcolor{cordHeader}
Labels & Treatment & RMSE & MAE & $R^2$ \\
\midrule
10\% & CORD (Scratch) & $0.1169 \pm 0.0059$ & $0.0959 \pm 0.0048$ & $0.8074 \pm 0.0191$ \\
\rowcolor{cordRow}
10\% & CORD (Single-domain) Frozen & $0.1245 \pm 0.0162$ & $0.1044 \pm 0.0155$ & $0.7793 \pm 0.0592$ \\
10\% & CORD (Multi-domain) Frozen & $\second{0.0864 \pm 0.0059}$ & $\second{0.0671 \pm 0.0060}$ & $\second{0.8948 \pm 0.0144}$ \\
\rowcolor{cordRow}
10\% & CORD (Single-domain) Partial FT & $0.1194 \pm 0.0105$ & $0.0996 \pm 0.0088$ & $0.7986 \pm 0.0348$ \\
10\% & CORD (Multi-domain) Partial FT & $\best{0.0791 \pm 0.0116}$ & $\best{0.0668 \pm 0.0107}$ & $\best{0.9105 \pm 0.0267}$ \\
\rowcolor{cordRow}
10\% & CORD (Single-domain) Full FT & $0.1082 \pm 0.0110$ & $0.0924 \pm 0.0107$ & $0.8340 \pm 0.0338$ \\
10\% & CORD (Multi-domain) Full FT & $0.0871 \pm 0.0245$ & $0.0761 \pm 0.0206$ & $0.8867 \pm 0.0638$ \\
\midrule
\rowcolor{cordRow}
20\% & CORD (Scratch) & $0.1007 \pm 0.0113$ & $0.0805 \pm 0.0114$ & $0.8561 \pm 0.0318$ \\
20\% & CORD (Single-domain) Frozen & $0.1131 \pm 0.0160$ & $0.0937 \pm 0.0150$ & $0.8173 \pm 0.0532$ \\
\rowcolor{cordRow}
20\% & CORD (Multi-domain) Frozen & $0.0782 \pm 0.0080$ & $0.0607 \pm 0.0074$ & $0.9132 \pm 0.0177$ \\
20\% & CORD (Single-domain) Partial FT & $0.1088 \pm 0.0106$ & $0.0903 \pm 0.0109$ & $0.8323 \pm 0.0322$ \\
\rowcolor{cordRow}
20\% & CORD (Multi-domain) Partial FT & $\second{0.0660 \pm 0.0049}$ & $\best{0.0514 \pm 0.0027}$ & $\second{0.9385 \pm 0.0092}$ \\
20\% & CORD (Single-domain) Full FT & $0.0930 \pm 0.0023$ & $0.0782 \pm 0.0046$ & $0.8785 \pm 0.0060$ \\
\rowcolor{cordRow}
20\% & CORD (Multi-domain) Full FT & $\best{0.0609 \pm 0.0096}$ & $\second{0.0529 \pm 0.0091}$ & $\best{0.9469 \pm 0.0152}$ \\
\midrule
100\% & CORD (Scratch) & $0.0782 \pm 0.0247$ & $0.0608 \pm 0.0197$ & $0.9073 \pm 0.0530$ \\
\rowcolor{cordRow}
100\% & CORD (Single-domain) Frozen & $0.1133 \pm 0.0056$ & $0.0911 \pm 0.0060$ & $0.8191 \pm 0.0180$ \\
100\% & CORD (Multi-domain) Frozen & $0.0658 \pm 0.0109$ & $0.0516 \pm 0.0102$ & $0.9378 \pm 0.0186$ \\
\rowcolor{cordRow}
100\% & CORD (Single-domain) Partial FT & $0.0920 \pm 0.0077$ & $0.0745 \pm 0.0065$ & $0.8805 \pm 0.0199$ \\
100\% & CORD (Multi-domain) Partial FT & $\best{0.0491 \pm 0.0079}$ & $\best{0.0413 \pm 0.0077}$ & $\best{0.9654 \pm 0.0111}$ \\
\rowcolor{cordRow}
100\% & CORD (Single-domain) Full FT & $0.0800 \pm 0.0216$ & $0.0671 \pm 0.0199$ & $0.9048 \pm 0.0436$ \\
100\% & CORD (Multi-domain) Full FT & $\second{0.0530 \pm 0.0165}$ & $\second{0.0436 \pm 0.0136}$ & $\second{0.9575 \pm 0.0250}$ \\
\bottomrule
\end{tabular}

\end{table}

\subsection{RUL prediction baselines}\label{app:baseline}
The baselines use model-appropriate observations and scalar RUL heads.
MOMENT is evaluated with a frozen encoder; the remaining baselines follow
their model-appropriate input processing and readout configurations.
\subsubsection{Bearing}
\begin{table}[H]\centering\small
\caption{Bearing RUL prediction comparison at 10\% labels (five seeds).}
\label{tab:appendix-baseline-bearing-10}
\begin{tabular}{llll}
\toprule
\rowcolor{cordHeader}
Method & RMSE & MAE & $R^2$ \\
\midrule
CORD (Scratch) & $0.1463 \pm 0.0393$ & $0.1172 \pm 0.0292$ & $0.7428 \pm 0.1406$ \\
\rowcolor{cordRow}
CORD (Single-domain) & $\best{0.1121 \pm 0.0224}$ & $\best{0.0833 \pm 0.0142}$ & $\best{0.8526 \pm 0.0623}$ \\
CORD (Multi-domain) & $\second{0.1240 \pm 0.0289}$ & $\second{0.0979 \pm 0.0239}$ & $\second{0.8177 \pm 0.0859}$ \\
\rowcolor{cordRow}
MLP & $0.2185 \pm 0.0179$ & $0.1992 \pm 0.0149$ & $0.4543 \pm 0.0919$ \\
Random Forest & $0.1840 \pm 0.0035$ & $0.1698 \pm 0.0032$ & $0.6148 \pm 0.0145$ \\
\rowcolor{cordRow}
XGBoost & $0.1494 \pm 0.0039$ & $0.1361 \pm 0.0054$ & $0.7462 \pm 0.0131$ \\
TCN & $0.2681 \pm 0.0286$ & $0.2186 \pm 0.0175$ & $0.1757 \pm 0.1767$ \\
\rowcolor{cordRow}
PatchTST & $0.3019 \pm 0.0367$ & $0.2547 \pm 0.0238$ & $-0.0482 \pm 0.2475$ \\
iTransformer & $0.3138 \pm 0.0207$ & $0.2513 \pm 0.0184$ & $-0.1232 \pm 0.1454$ \\
\rowcolor{cordRow}
MOMENT & $0.3070 \pm 0.0018$ & $0.2696 \pm 0.0021$ & $-0.0714 \pm 0.0128$ \\
\bottomrule
\end{tabular}

\end{table}
\begin{table}[H]\centering\small
\caption{Bearing RUL prediction comparison at 20\% labels (five seeds).}
\label{tab:appendix-baseline-bearing-20}
\begin{tabular}{llll}
\toprule
\rowcolor{cordHeader}
Method & RMSE & MAE & $R^2$ \\
\midrule
CORD (Scratch) & $\second{0.1400 \pm 0.0145}$ & $\second{0.1081 \pm 0.0112}$ & $\second{0.7752 \pm 0.0453}$ \\
\rowcolor{cordRow}
CORD (Single-domain) & $0.1464 \pm 0.0142$ & $0.1100 \pm 0.0132$ & $0.7546 \pm 0.0474$ \\
CORD (Multi-domain) & $\best{0.1329 \pm 0.0214}$ & $\best{0.1044 \pm 0.0142}$ & $\best{0.7951 \pm 0.0697}$ \\
\rowcolor{cordRow}
MLP & $0.2307 \pm 0.0124$ & $0.2093 \pm 0.0106$ & $0.3937 \pm 0.0644$ \\
Random Forest & $0.1910 \pm 0.0022$ & $0.1664 \pm 0.0017$ & $0.5853 \pm 0.0096$ \\
\rowcolor{cordRow}
XGBoost & $0.1824 \pm 0.0043$ & $0.1548 \pm 0.0060$ & $0.6218 \pm 0.0176$ \\
TCN & $0.2930 \pm 0.0477$ & $0.2323 \pm 0.0324$ & $0.0035 \pm 0.3391$ \\
\rowcolor{cordRow}
PatchTST & $0.3338 \pm 0.0080$ & $0.2736 \pm 0.0078$ & $-0.2670 \pm 0.0607$ \\
iTransformer & $0.3157 \pm 0.0280$ & $0.2611 \pm 0.0314$ & $-0.1404 \pm 0.2036$ \\
\rowcolor{cordRow}
MOMENT & $0.3176 \pm 0.0017$ & $0.2761 \pm 0.0014$ & $-0.1471 \pm 0.0120$ \\
\bottomrule
\end{tabular}

\end{table}
\begin{table}[H]\centering\small
\caption{Bearing RUL prediction comparison at 100\% labels (five seeds).}
\label{tab:appendix-baseline-bearing-100}
\begin{tabular}{llll}
\toprule
\rowcolor{cordHeader}
Method & RMSE & MAE & $R^2$ \\
\midrule
CORD (Scratch) & $0.1122 \pm 0.0233$ & $0.0854 \pm 0.0143$ & $0.8519 \pm 0.0637$ \\
\rowcolor{cordRow}
CORD (Single-domain) & $\second{0.0988 \pm 0.0184}$ & $\best{0.0718 \pm 0.0162}$ & $\second{0.8859 \pm 0.0436}$ \\
CORD (Multi-domain) & $\best{0.0972 \pm 0.0071}$ & $\second{0.0744 \pm 0.0076}$ & $\best{0.8921 \pm 0.0160}$ \\
\rowcolor{cordRow}
MLP & $0.2160 \pm 0.0114$ & $0.1922 \pm 0.0100$ & $0.4685 \pm 0.0553$ \\
Random Forest & $0.1853 \pm 0.0009$ & $0.1570 \pm 0.0009$ & $0.6095 \pm 0.0040$ \\
\rowcolor{cordRow}
XGBoost & $0.1846 \pm 0.0031$ & $0.1619 \pm 0.0032$ & $0.6125 \pm 0.0130$ \\
TCN & $0.2287 \pm 0.0073$ & $0.1893 \pm 0.0058$ & $0.4049 \pm 0.0389$ \\
\rowcolor{cordRow}
PatchTST & $0.2663 \pm 0.0243$ & $0.2230 \pm 0.0193$ & $0.1882 \pm 0.1497$ \\
iTransformer & $0.3071 \pm 0.0281$ & $0.2480 \pm 0.0200$ & $-0.0796 \pm 0.1996$ \\
\rowcolor{cordRow}
MOMENT & $0.2711 \pm 0.0023$ & $0.2238 \pm 0.0020$ & $0.1646 \pm 0.0142$ \\
\bottomrule
\end{tabular}

\end{table}
\subsubsection{Battery}
\begin{table}[H]\centering\small
\caption{Battery RUL prediction comparison at 10\% labels (five seeds).}
\label{tab:appendix-baseline-battery-10}
\begin{tabular}{llll}
\toprule
\rowcolor{cordHeader}
Method & RMSE & MAE & $R^2$ \\
\midrule
CORD (Scratch) & $0.0711 \pm 0.0091$ & $0.0516 \pm 0.0065$ & $0.9384 \pm 0.0146$ \\
\rowcolor{cordRow}
CORD (Single-domain) & $0.0689 \pm 0.0048$ & $\second{0.0492 \pm 0.0035}$ & $0.9426 \pm 0.0078$ \\
CORD (Multi-domain) & $\best{0.0673 \pm 0.0064}$ & $\best{0.0487 \pm 0.0044}$ & $\best{0.9450 \pm 0.0105}$ \\
\rowcolor{cordRow}
MLP & $0.1526 \pm 0.0130$ & $0.1026 \pm 0.0071$ & $0.7178 \pm 0.0469$ \\
Random Forest & $0.1085 \pm 0.0006$ & $0.0723 \pm 0.0003$ & $0.8581 \pm 0.0015$ \\
\rowcolor{cordRow}
XGBoost & $0.1036 \pm 0.0030$ & $0.0716 \pm 0.0021$ & $0.8705 \pm 0.0074$ \\
TCN & $0.0807 \pm 0.0068$ & $0.0623 \pm 0.0062$ & $0.9210 \pm 0.0129$ \\
\rowcolor{cordRow}
PatchTST & $\second{0.0676 \pm 0.0080}$ & $0.0535 \pm 0.0070$ & $\second{0.9443 \pm 0.0123}$ \\
iTransformer & $0.0904 \pm 0.0018$ & $0.0684 \pm 0.0011$ & $0.9015 \pm 0.0039$ \\
\rowcolor{cordRow}
MOMENT & $0.0730 \pm 0.0021$ & $0.0594 \pm 0.0017$ & $0.9358 \pm 0.0038$ \\
\bottomrule
\end{tabular}

\end{table}
\begin{table}[H]\centering\small
\caption{Battery RUL prediction comparison at 20\% labels (five seeds).}
\label{tab:appendix-baseline-battery-20}
\begin{tabular}{llll}
\toprule
\rowcolor{cordHeader}
Method & RMSE & MAE & $R^2$ \\
\midrule
CORD (Scratch) & $\best{0.0680 \pm 0.0060}$ & $\second{0.0497 \pm 0.0056}$ & $\best{0.9439 \pm 0.0101}$ \\
\rowcolor{cordRow}
CORD (Single-domain) & $0.0735 \pm 0.0050$ & $0.0533 \pm 0.0044$ & $0.9346 \pm 0.0089$ \\
CORD (Multi-domain) & $\second{0.0683 \pm 0.0043}$ & $\best{0.0484 \pm 0.0035}$ & $\second{0.9436 \pm 0.0072}$ \\
\rowcolor{cordRow}
MLP & $0.1408 \pm 0.0081$ & $0.0968 \pm 0.0041$ & $0.7605 \pm 0.0275$ \\
Random Forest & $0.1110 \pm 0.0011$ & $0.0730 \pm 0.0006$ & $0.8516 \pm 0.0030$ \\
\rowcolor{cordRow}
XGBoost & $0.1052 \pm 0.0009$ & $0.0722 \pm 0.0005$ & $0.8668 \pm 0.0023$ \\
TCN & $0.0813 \pm 0.0064$ & $0.0634 \pm 0.0055$ & $0.9199 \pm 0.0124$ \\
\rowcolor{cordRow}
PatchTST & $0.0722 \pm 0.0062$ & $0.0548 \pm 0.0063$ & $0.9368 \pm 0.0105$ \\
iTransformer & $0.0893 \pm 0.0044$ & $0.0685 \pm 0.0043$ & $0.9037 \pm 0.0096$ \\
\rowcolor{cordRow}
MOMENT & $0.0695 \pm 0.0013$ & $0.0556 \pm 0.0013$ & $0.9418 \pm 0.0022$ \\
\bottomrule
\end{tabular}

\end{table}
\begin{table}[H]\centering\small
\caption{Battery RUL prediction comparison at 100\% labels (five seeds).}
\label{tab:appendix-baseline-battery-100}
\begin{tabular}{llll}
\toprule
\rowcolor{cordHeader}
Method & RMSE & MAE & $R^2$ \\
\midrule
CORD (Scratch) & $\best{0.0597 \pm 0.0085}$ & $\best{0.0435 \pm 0.0073}$ & $\best{0.9564 \pm 0.0125}$ \\
\rowcolor{cordRow}
CORD (Single-domain) & $0.0652 \pm 0.0064$ & $0.0464 \pm 0.0054$ & $0.9483 \pm 0.0103$ \\
CORD (Multi-domain) & $0.0632 \pm 0.0056$ & $\second{0.0458 \pm 0.0050}$ & $0.9515 \pm 0.0086$ \\
\rowcolor{cordRow}
MLP & $0.1005 \pm 0.0052$ & $0.0769 \pm 0.0032$ & $0.8780 \pm 0.0126$ \\
Random Forest & $0.1054 \pm 0.0002$ & $0.0712 \pm 0.0001$ & $0.8662 \pm 0.0005$ \\
\rowcolor{cordRow}
XGBoost & $0.0992 \pm 0.0013$ & $0.0696 \pm 0.0011$ & $0.8814 \pm 0.0031$ \\
TCN & $0.0788 \pm 0.0053$ & $0.0622 \pm 0.0043$ & $0.9250 \pm 0.0101$ \\
\rowcolor{cordRow}
PatchTST & $0.0792 \pm 0.0049$ & $0.0592 \pm 0.0041$ & $0.9242 \pm 0.0094$ \\
iTransformer & $0.0813 \pm 0.0008$ & $0.0615 \pm 0.0008$ & $0.9203 \pm 0.0016$ \\
\rowcolor{cordRow}
MOMENT & $\second{0.0620 \pm 0.0028}$ & $0.0485 \pm 0.0028$ & $\second{0.9536 \pm 0.0042}$ \\
\bottomrule
\end{tabular}

\end{table}
\subsubsection{Cutting tools}
\begin{table}[H]\centering\small
\caption{Cutting-tool RUL prediction comparison at 10\% labels (five seeds).}
\label{tab:appendix-baseline-milling-10}
\begin{tabular}{llll}
\toprule
\rowcolor{cordHeader}
Method & RMSE & MAE & $R^2$ \\
\midrule
CORD (Scratch) & $0.1169 \pm 0.0059$ & $0.0959 \pm 0.0048$ & $0.8074 \pm 0.0191$ \\
\rowcolor{cordRow}
CORD (Single-domain) & $0.1082 \pm 0.0110$ & $0.0924 \pm 0.0107$ & $0.8340 \pm 0.0338$ \\
CORD (Multi-domain) & $\best{0.0871 \pm 0.0245}$ & $\best{0.0761 \pm 0.0206}$ & $\best{0.8867 \pm 0.0638}$ \\
\rowcolor{cordRow}
MLP & $\second{0.0991 \pm 0.0161}$ & $\second{0.0792 \pm 0.0152}$ & $\second{0.8591 \pm 0.0475}$ \\
Random Forest & $0.1117 \pm 0.0034$ & $0.0922 \pm 0.0045$ & $0.8244 \pm 0.0107$ \\
\rowcolor{cordRow}
XGBoost & $0.1258 \pm 0.0023$ & $0.1021 \pm 0.0021$ & $0.7775 \pm 0.0081$ \\
TCN & $0.1409 \pm 0.0270$ & $0.1123 \pm 0.0221$ & $0.7128 \pm 0.1080$ \\
\rowcolor{cordRow}
PatchTST & $0.2258 \pm 0.0130$ & $0.1833 \pm 0.0090$ & $0.2818 \pm 0.0828$ \\
iTransformer & $0.2728 \pm 0.0085$ & $0.2268 \pm 0.0085$ & $-0.0465 \pm 0.0650$ \\
\rowcolor{cordRow}
MOMENT & $0.2452 \pm 0.0052$ & $0.2023 \pm 0.0039$ & $0.1548 \pm 0.0358$ \\
\bottomrule
\end{tabular}

\end{table}
\begin{table}[H]\centering\small
\caption{Cutting-tool RUL prediction comparison at 20\% labels (five seeds).}
\label{tab:appendix-baseline-milling-20}
\begin{tabular}{llll}
\toprule
\rowcolor{cordHeader}
Method & RMSE & MAE & $R^2$ \\
\midrule
CORD (Scratch) & $0.1007 \pm 0.0113$ & $0.0805 \pm 0.0114$ & $0.8561 \pm 0.0318$ \\
\rowcolor{cordRow}
CORD (Single-domain) & $\second{0.0930 \pm 0.0023}$ & $\second{0.0782 \pm 0.0046}$ & $\second{0.8785 \pm 0.0060}$ \\
CORD (Multi-domain) & $\best{0.0609 \pm 0.0096}$ & $\best{0.0529 \pm 0.0091}$ & $\best{0.9469 \pm 0.0152}$ \\
\rowcolor{cordRow}
MLP & $0.1085 \pm 0.0132$ & $0.0862 \pm 0.0104$ & $0.8325 \pm 0.0414$ \\
Random Forest & $0.1202 \pm 0.0024$ & $0.0999 \pm 0.0022$ & $0.7968 \pm 0.0079$ \\
\rowcolor{cordRow}
XGBoost & $0.1204 \pm 0.0012$ & $0.1010 \pm 0.0018$ & $0.7962 \pm 0.0040$ \\
TCN & $0.1266 \pm 0.0117$ & $0.1017 \pm 0.0090$ & $0.7730 \pm 0.0413$ \\
\rowcolor{cordRow}
PatchTST & $0.2103 \pm 0.0086$ & $0.1733 \pm 0.0057$ & $0.3775 \pm 0.0507$ \\
iTransformer & $0.2582 \pm 0.0085$ & $0.2137 \pm 0.0080$ & $0.0626 \pm 0.0629$ \\
\rowcolor{cordRow}
MOMENT & $0.2431 \pm 0.0080$ & $0.1989 \pm 0.0055$ & $0.1689 \pm 0.0552$ \\
\bottomrule
\end{tabular}

\end{table}
\begin{table}[H]\centering\small
\caption{Cutting-tool RUL prediction comparison at 100\% labels (five seeds).}
\label{tab:appendix-baseline-milling-100}
\begin{tabular}{llll}
\toprule
\rowcolor{cordHeader}
Method & RMSE & MAE & $R^2$ \\
\midrule
CORD (Scratch) & $0.0782 \pm 0.0247$ & $0.0608 \pm 0.0197$ & $0.9073 \pm 0.0530$ \\
\rowcolor{cordRow}
CORD (Single-domain) & $0.0800 \pm 0.0216$ & $0.0671 \pm 0.0199$ & $0.9048 \pm 0.0436$ \\
CORD (Multi-domain) & $\best{0.0530 \pm 0.0165}$ & $\best{0.0436 \pm 0.0136}$ & $\best{0.9575 \pm 0.0250}$ \\
\rowcolor{cordRow}
MLP & $0.1177 \pm 0.0128$ & $0.0794 \pm 0.0077$ & $0.8034 \pm 0.0449$ \\
Random Forest & $0.1230 \pm 0.0014$ & $0.1014 \pm 0.0022$ & $0.7874 \pm 0.0050$ \\
\rowcolor{cordRow}
XGBoost & $0.1214 \pm 0.0016$ & $0.0992 \pm 0.0011$ & $0.7927 \pm 0.0056$ \\
TCN & $\second{0.0722 \pm 0.0070}$ & $\second{0.0581 \pm 0.0057}$ & $\second{0.9261 \pm 0.0144}$ \\
\rowcolor{cordRow}
PatchTST & $0.1810 \pm 0.0158$ & $0.1485 \pm 0.0103$ & $0.5368 \pm 0.0806$ \\
iTransformer & $0.1890 \pm 0.0038$ & $0.1657 \pm 0.0038$ & $0.4977 \pm 0.0204$ \\
\rowcolor{cordRow}
MOMENT & $0.2421 \pm 0.0022$ & $0.2121 \pm 0.0036$ & $0.1763 \pm 0.0151$ \\
\bottomrule
\end{tabular}

\end{table}

\section{Generalization under Pretraining-Excluded System Types: Engine Adaptation}\label{app:engine}
The engine comparison isolates initialization. Both conditions use the same
observation-interface construction, engine interface, seven-flight context, MSE objective,
optimizer, label-budget samples, and 400 supervised optimizer updates. Scratch
starts all parameters randomly; Source-pretrained initializes the reusable
encoder from the selected bearing--battery--cutting-tool checkpoint and adapts
all trainable components. The 400-update adaptation budget is selected by
grouped cross-validation on U2/U5/U10/U16/U18/U20 and then fixed for all final
U11 runs. Tables report mean $\pm$ sample SD for seeds 42--46. Within each
label budget, \best{red bold} marks the better mean and
\second{blue bold underlined} marks the second-best mean.
\begin{table}[H]\centering
\caption{N-CMAPSS U11, 10\% labels: matched Scratch and source-pretrained initialization (five seeds).}
\label{tab:appendix-engine-10}
\small\begin{tabular}{lllll}
\toprule
\rowcolor{cordHeader}
Initialization & Updates & RMSE & MAE & $R^2$ \\
\midrule
CORD (Scratch) & 400 & $\second{0.1463 \pm 0.0134}$ & $\second{0.1151 \pm 0.0103}$ & $\second{0.6518 \pm 0.0611}$ \\
\rowcolor{cordRow}
CORD (Source-pretrained) & 400 & $\best{0.1374 \pm 0.0117}$ & $\best{0.1098 \pm 0.0091}$ & $\best{0.6932 \pm 0.0510}$ \\
\bottomrule
\end{tabular}

\end{table}
\begin{table}[H]\centering
\caption{N-CMAPSS U11, 20\% labels: matched Scratch and source-pretrained initialization (five seeds).}
\label{tab:appendix-engine-20}
\small\begin{tabular}{lllll}
\toprule
\rowcolor{cordHeader}
Initialization & Updates & RMSE & MAE & $R^2$ \\
\midrule
CORD (Scratch) & 400 & $\second{0.1163 \pm 0.0131}$ & $\second{0.0903 \pm 0.0090}$ & $\second{0.7795 \pm 0.0482}$ \\
\rowcolor{cordRow}
CORD (Source-pretrained) & 400 & $\best{0.1122 \pm 0.0069}$ & $\best{0.0883 \pm 0.0070}$ & $\best{0.7959 \pm 0.0252}$ \\
\bottomrule
\end{tabular}

\end{table}
\begin{table}[H]\centering
\caption{N-CMAPSS U11, 100\% labels: matched Scratch and source-pretrained initialization (five seeds).}
\label{tab:appendix-engine-100}
\small\begin{tabular}{lllll}
\toprule
\rowcolor{cordHeader}
Initialization & Updates & RMSE & MAE & $R^2$ \\
\midrule
CORD (Scratch) & 400 & $\best{0.0768 \pm 0.0154}$ & $\best{0.0566 \pm 0.0112}$ & $\best{0.9018 \pm 0.0390}$ \\
\rowcolor{cordRow}
CORD (Source-pretrained) & 400 & $\second{0.0841 \pm 0.0099}$ & $\second{0.0582 \pm 0.0055}$ & $\second{0.8844 \pm 0.0274}$ \\
\bottomrule
\end{tabular}

\end{table}

Table~\ref{tab:appendix-idm-objective} isolates IDM under the matched protocol
of Section~\ref{sec:ablation}; architecture and downstream adaptation are fixed.
\begin{table}[H]\centering\footnotesize
\caption{IDM pretraining ablation on N-CMAPSS U11 (five seeds; 400 adaptation updates).}
\label{tab:appendix-idm-objective}
\begin{tabular}{llccc}
\toprule
\rowcolor{cordHeader}
Labels & Pretraining objective & RMSE $\downarrow$ & MAE $\downarrow$ & $R^2\uparrow$ \\
\midrule
10\% & ISM only & $\second{0.17251\pm0.01562}$ & $\second{0.14087\pm0.01404}$ & $\second{0.51609\pm0.08495}$ \\
\rowcolor{cordRow}
10\% & ISM + IDM & $\best{0.13742\pm0.01169}$ & $\best{0.10979\pm0.00914}$ & $\best{0.69319\pm0.05103}$ \\
\midrule
20\% & ISM only & $\second{0.14477\pm0.00588}$ & $\second{0.11872\pm0.00390}$ & $\second{0.66099\pm0.02775}$ \\
\rowcolor{cordRow}
20\% & ISM + IDM & $\best{0.11225\pm0.00686}$ & $\best{0.08833\pm0.00698}$ & $\best{0.79586\pm0.02515}$ \\
\midrule
100\% & ISM only & $\second{0.09439\pm0.01887}$ & $\second{0.07094\pm0.01156}$ & $\second{0.85148\pm0.06445}$ \\
\rowcolor{cordRow}
100\% & ISM + IDM & $\best{0.08415\pm0.00987}$ & $\best{0.05819\pm0.00546}$ & $\best{0.88436\pm0.02740}$ \\
\bottomrule
\end{tabular}

\end{table}

\begin{figure}[H]
\centering\includegraphics[width=.52\linewidth]{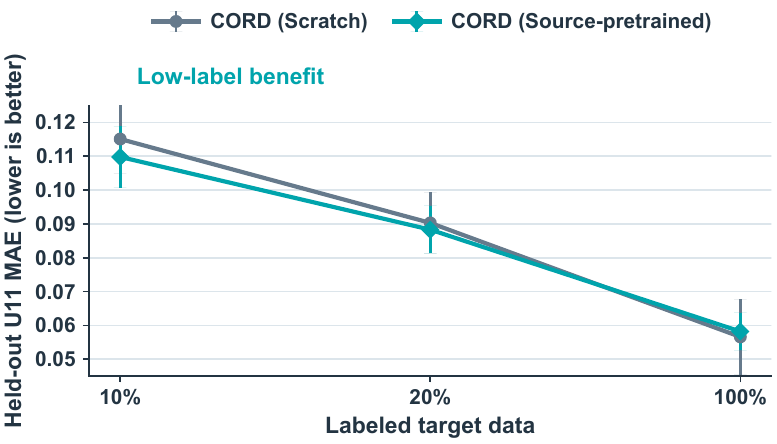}
\caption{Engine MAE under the same matched 400-update protocol as
Figure~\ref{fig:engine}. Points and error bars show mean $\pm$
one sample SD over five seeds.}
\label{fig:engine-mae}
\end{figure}
\begin{figure}[H]
\centering\includegraphics[width=.52\linewidth]{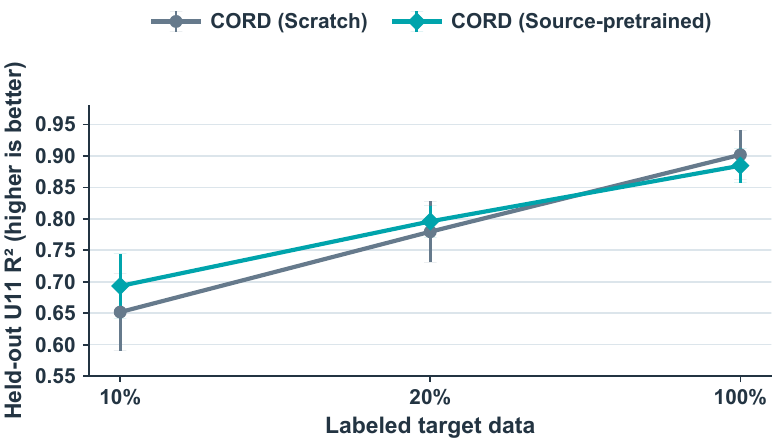}
\caption{Engine $R^2$ under the same matched 400-update protocol as
Figure~\ref{fig:engine}. Points and error bars show mean $\pm$
sample SD over five seeds.}
\label{fig:engine-r2}
\end{figure}

\section{Frozen-Representation Reuse}\label{app:representation}
Frozen-encoder analysis uses one selected checkpoint per condition. Held-out
unit queries retrieve five nearest reference-unit states, and normalized-RUL
discrepancy measures lifecycle-neighborhood quality. Bearings and batteries use
the 96-dimensional projector and cutting tools the 192-dimensional final-layer
normalized hidden representation.
\begin{table}[H]
\centering\small
\caption{Cross-unit 5-NN normalized-RUL error on held-out units. Lower is
better. Within each system type, \best{red bold} marks the best encoder and
\second{blue bold underlined} the second-best. Multi-domain pretraining
improves over Single-domain for all three system types.}
\label{tab:appendix-geometry}
\begin{tabular}{lll}
\toprule
\rowcolor{cordHeader}
System type & Encoder & 5-NN normalized-RUL error \\
\midrule
Bearings & CORD (Random init.) & 0.2585 \\
\rowcolor{cordRow}
Bearings & CORD (Single-domain) & \second{0.1407} \\
Bearings & CORD (Multi-domain) & \best{0.1212} \\
\rowcolor{cordRow}
Batteries & CORD (Random init.) & 0.0793 \\
Batteries & CORD (Single-domain) & \second{0.0736} \\
\rowcolor{cordRow}
Batteries & CORD (Multi-domain) & \best{0.0604} \\
Cutting tools & CORD (Random init.) & \second{0.1054} \\
\rowcolor{cordRow}
Cutting tools & CORD (Single-domain) & 0.1376 \\
Cutting tools & CORD (Multi-domain) & \best{0.1000} \\
\bottomrule
\end{tabular}

\end{table}

Figure~\ref{fig:pca} visualizes lifecycle organization within the same selected
frozen encoder conditions.

\section{Observation Interface: Structured Descriptors versus Raw-Resampled Inputs}\label{app:raw26}
This scratch-only comparison uses the same FP32 downstream split and fitting
protocol at 10\%, 20\%, and 100\% labels, with seeds 42--46. The
Structured-Descriptor condition uses the same Scratch results reported above,
whereas
the Raw-Resampled condition replaces descriptor extraction with parameter-free
resampling of native signal windows to 26 values while retaining the same
downstream split and training protocol. Both use 65-token, width-96, two-layer
encoders and the corresponding type-specific RUL readouts. The displayed
uncertainty is sample SD over downstream seeds on one fixed test
unit per system type. In all nine settings, Structured Descriptors are better
on all three metrics; they also win all 45 matched-seed comparisons for each
metric. The advantage of Structured Descriptors persists at the 100\% label budget.
\begin{table}[H]
\centering
\caption{Structured Descriptors versus Raw-Resampled Inputs under the matched Scratch protocol: mean $\pm$ SD over five seeds. Red bold marks the better input within each system type and label budget; blue underlined marks the other.}
\label{tab:raw26}
\begingroup\small\setlength{\tabcolsep}{2.5pt}
% Source: code/healthtoken_final_study/B_ablation_study/02_raw26_e38_matched/results/summary.json
% E38 FP32 scratch protocol; five downstream seeds (42--46); sample SD.
\begin{tabular}{@{}lllrrr@{}}
\toprule
\rowcolor{cordHeader}
System type & Labels & Input representation & RMSE $\downarrow$ & MAE $\downarrow$ & $R^2\uparrow$ \\
\midrule
Bearings & 10\% & Structured Descriptors & $\best{0.1463\pm0.0393}$ & $\best{0.1172\pm0.0292}$ & $\best{0.7428\pm0.1406}$ \\
\rowcolor{cordRow}
 & & Raw-Resampled Inputs & $\second{0.3590\pm0.0237}$ & $\second{0.3016\pm0.0212}$ & $\second{-0.4704\pm0.1947}$ \\
Bearings & 20\% & Structured Descriptors & $\best{0.1400\pm0.0145}$ & $\best{0.1081\pm0.0112}$ & $\best{0.7752\pm0.0453}$ \\
\rowcolor{cordRow}
 & & Raw-Resampled Inputs & $\second{0.3157\pm0.0195}$ & $\second{0.2691\pm0.0131}$ & $\second{-0.1362\pm0.1428}$ \\
Bearings & 100\% & Structured Descriptors & $\best{0.1122\pm0.0233}$ & $\best{0.0854\pm0.0143}$ & $\best{0.8519\pm0.0637}$ \\
\rowcolor{cordRow}
 & & Raw-Resampled Inputs & $\second{0.3091\pm0.0338}$ & $\second{0.2619\pm0.0283}$ & $\second{-0.0966\pm0.2323}$ \\
\midrule
Batteries & 10\% & Structured Descriptors & $\best{0.0711\pm0.0091}$ & $\best{0.0516\pm0.0065}$ & $\best{0.9384\pm0.0146}$ \\
\rowcolor{cordRow}
 & & Raw-Resampled Inputs & $\second{0.2131\pm0.0112}$ & $\second{0.1664\pm0.0133}$ & $\second{0.4519\pm0.0579}$ \\
Batteries & 20\% & Structured Descriptors & $\best{0.0680\pm0.0060}$ & $\best{0.0497\pm0.0056}$ & $\best{0.9439\pm0.0101}$ \\
\rowcolor{cordRow}
 & & Raw-Resampled Inputs & $\second{0.1715\pm0.0065}$ & $\second{0.1213\pm0.0092}$ & $\second{0.6454\pm0.0269}$ \\
Batteries & 100\% & Structured Descriptors & $\best{0.0597\pm0.0085}$ & $\best{0.0435\pm0.0073}$ & $\best{0.9564\pm0.0125}$ \\
\rowcolor{cordRow}
 & & Raw-Resampled Inputs & $\second{0.1724\pm0.0113}$ & $\second{0.1097\pm0.0070}$ & $\second{0.6408\pm0.0463}$ \\
\midrule
Cutting tools & 10\% & Structured Descriptors & $\best{0.1169\pm0.0059}$ & $\best{0.0959\pm0.0048}$ & $\best{0.8074\pm0.0191}$ \\
\rowcolor{cordRow}
 & & Raw-Resampled Inputs & $\second{0.2845\pm0.0223}$ & $\second{0.2363\pm0.0152}$ & $\second{-0.1435\pm0.1828}$ \\
Cutting tools & 20\% & Structured Descriptors & $\best{0.1007\pm0.0113}$ & $\best{0.0805\pm0.0114}$ & $\best{0.8561\pm0.0318}$ \\
\rowcolor{cordRow}
 & & Raw-Resampled Inputs & $\second{0.2178\pm0.0421}$ & $\second{0.1755\pm0.0318}$ & $\second{0.3132\pm0.2656}$ \\
Cutting tools & 100\% & Structured Descriptors & $\best{0.0782\pm0.0247}$ & $\best{0.0608\pm0.0197}$ & $\best{0.9073\pm0.0530}$ \\
\rowcolor{cordRow}
 & & Raw-Resampled Inputs & $\second{0.1894\pm0.0254}$ & $\second{0.1531\pm0.0235}$ & $\second{0.4885\pm0.1421}$ \\
\bottomrule
\end{tabular}

\endgroup
\end{table}

The Structured-Descriptor condition uses the proposed local and global
observation descriptors. Bearings and cutting tools use 26 statistical and
spectral coordinates: mean, absolute mean, standard deviation, variance, RMS,
energy, peak, peak-to-peak range, minimum, maximum, skewness, kurtosis, crest,
shape, impulse, and clearance factors, root amplitude, zero-crossing rate,
spectral centroid, spectral bandwidth, spectral entropy, dominant frequency,
low-, mid-, and high-band energy, and the high-to-low-band energy ratio.
Batteries use voltage mean, standard deviation, minimum, maximum, range,
skewness, kurtosis, slope, mean absolute derivative, and mean absolute
curvature; C-rate mean, standard deviation, minimum, maximum, mean absolute
value, and slope; temperature mean, standard deviation, minimum, maximum,
change, and slope; local duration, capacity, and energy; and mean $dV/dQ$.
Global descriptors use the corresponding coordinates over the complete
observation extent, with full-cycle duration and capacity used for batteries
when available.

\section{Computational Cost and Shared Deployment}\label{app:efficiency}
\subsection{Frozen shared deployment}
\begin{table}[H]
\centering\small
\caption{Parameter storage for one frozen shared encoder with three task heads versus three separate extracted encoders.}
\label{tab:appendix-shared-deployment}
\begin{tabular}{llll}
\toprule
\rowcolor{cordHeader}
Deployment & Parameters & FP32 MiB & Relative \\
\midrule
One frozen shared encoder + three heads & 657,235 & 2.507 & 35.75\% reduction \\
\rowcolor{cordRow}
Three extracted encoders + three heads & 1,022,899 & 3.902 & Reference \\
\bottomrule
\end{tabular}

\end{table}
The parameter reduction applies when one frozen encoding system serves three
specialized heads. Independently fine-tuned encoders diverge and require
separate copies. The 35.75\% reduction refers to parameter storage under frozen
shared deployment; latency is reported separately in
Table~\ref{tab:appendix-efficiency}.

\subsection{A100 benchmark}
The A100-SXM4-40GB benchmark uses BF16 autocast with FP32 parameters,
resident model/input tensors, five warm-up calls, and five measurement rounds.
These rounds are timing repetitions, not accuracy seeds. CPU tree-model
measurements are excluded from the GPU comparison. Raw-signal preprocessing
and total source-pretraining GPU-hours are not included. These costs have
different directions and scopes, so no single best/second-best ranking is
assigned to the efficiency table. Table~\ref{tab:appendix-efficiency} reports
latency, throughput, memory, and update time as separate efficiency dimensions.
\begin{table}[H]
\centering
\caption{A100 inference and training efficiency by system type and predictor. B1 denotes batch-one latency; B32 denotes batch-32 throughput and peak memory.}
\label{tab:appendix-efficiency}
\begingroup\small
\begin{tabular}{lllllllll}
\toprule
\rowcolor{cordHeader}
System type & Model & Ch. & Parameters & Trainable & B1 ms & B32 samples/s & B32 MiB & Update s \\
\midrule
Bearings & CORD Full & 2 & 279,209 & 279,209 & 6.275 & 3,855 & 73 & 0.338 \\
\rowcolor{cordRow}
Bearings & CORD Partial & 2 & 279,209 & 161,145 & 6.332 & 3,821 & 73 & 0.261 \\
Bearings & CORD Frozen & 2 & 279,209 & 81,249 & 6.349 & 3,828 & 73 & 0.198 \\
\rowcolor{cordRow}
Bearings & TCN & 2 & 85,505 & 85,505 & 0.779 & 41,662 & 59 & 0.052 \\
Bearings & PatchTST & 2 & 231,649 & 231,649 & 2.804 & 11,029 & 38 & 0.166 \\
\rowcolor{cordRow}
Bearings & iTransformer & 2 & 245,953 & 245,953 & 2.876 & 10,643 & 38 & 0.172 \\
Bearings & MOMENT & 2 & 109,656,961 & 21,505 & 19.554 & 1,648 & 634 & 0.350 \\
\rowcolor{cordRow}
Batteries & CORD Full & 1 & 376,137 & 376,137 & 6.614 & 4,702 & 160 & 0.356 \\
Batteries & CORD Partial & 1 & 376,137 & 258,073 & 6.699 & 4,623 & 160 & 0.271 \\
\rowcolor{cordRow}
Batteries & CORD Frozen & 1 & 376,137 & 178,177 & 6.815 & 4,568 & 160 & 0.210 \\
Batteries & TCN & 1 & 85,057 & 85,057 & 0.761 & 42,697 & 118 & 0.053 \\
\rowcolor{cordRow}
Batteries & PatchTST & 1 & 234,529 & 234,529 & 2.738 & 10,891 & 41 & 0.168 \\
Batteries & iTransformer & 1 & 289,153 & 289,153 & 2.840 & 10,856 & 39 & 0.168 \\
\rowcolor{cordRow}
Batteries & MOMENT & 1 & 109,658,497 & 23,041 & 19.275 & 1,637 & 557 & 0.354 \\
Cutting tools & CORD Full & 7 & 367,553 & 367,553 & 5.627 & 4,990 & 224 & 0.291 \\
\rowcolor{cordRow}
Cutting tools & CORD Partial & 7 & 367,553 & 249,489 & 5.588 & 4,914 & 223 & 0.211 \\
Cutting tools & CORD Frozen & 7 & 367,553 & 169,593 & 5.919 & 4,853 & 223 & 0.168 \\
\rowcolor{cordRow}
Cutting tools & TCN & 7 & 88,385 & 88,385 & 0.766 & 34,605 & 143 & 0.052 \\
Cutting tools & PatchTST & 7 & 234,337 & 234,337 & 2.983 & 9,088 & 91 & 0.171 \\
\rowcolor{cordRow}
Cutting tools & iTransformer & 7 & 288,961 & 288,961 & 2.902 & 9,848 & 92 & 0.172 \\
Cutting tools & MOMENT & 7 & 109,656,961 & 21,505 & 19.484 & 537 & 1068 & 0.351 \\
\bottomrule
\end{tabular}

\endgroup
\end{table}

\end{document}